\documentclass[12pt]{article}
\usepackage[english]{babel}
\usepackage[a4paper,top=2.5cm,bottom=2.5cm,left=2.5cm,right=2.5cm,marginparwidth=1.75cm]{geometry}
\usepackage{amsmath}
\usepackage{amssymb}
\usepackage{graphicx}
\usepackage[colorlinks=true, allcolors=blue]{hyperref}
\usepackage{comment}
\usepackage[font=small]{caption}
\usepackage{subcaption}
\newcommand{\textttlb}[1]{%
  {\ttfamily\def\_{\textunderscore\allowbreak}#1}%
}

\usepackage{booktabs}
\usepackage{listings}
\usepackage{xcolor}
\definecolor{codegray}{rgb}{0.5,0.5,0.5}
\definecolor{backcolor}{rgb}{0.95,0.95,0.92}
\lstdefinestyle{mypython}{
    backgroundcolor=\color{backcolor},   
    commentstyle=\color{codegray},
    keywordstyle=\color{blue},
    numberstyle=\tiny\color{gray},
    stringstyle=\color{red},
    basicstyle=\ttfamily\footnotesize,
    breaklines=true,
    captionpos=b,
    keepspaces=true,
    numbers=left,
    numbersep=5pt,
    showspaces=false,
    showstringspaces=false,
    showtabs=false,
    tabsize=4,
    language=Python
}
\newcommand{\keycluster}{\mathcal{K}}
\newcommand{\querycluster}{\mathcal{Q}}

\definecolor{adcolor}{rgb}{0.6,0,0.21}
\definecolor{kncolor}{rgb}{0.1,0.6,0.1}

\RequirePackage[backend=biber,sorting=none,citestyle=numeric,style=numeric, doi=false, isbn=false, url=false]{biblatex}
\title{ClusterAttention: A training-free speedup of bidirectional attention}
\author{Kasper Nordenram\textsuperscript{*} \and Amelie Dittmann\textsuperscript{*}}
\date{\small\textsuperscript{*}Independent researcher \vspace{1em} \\ 
\today
}

\begin{document}
\maketitle

\begin{abstract}
    We introduce ClusterAttention, a general training-free speedup of bidirectional attention at large token counts. 
    We point out two common assumptions in contemporary training-free methods; attention sparsity, and context that can be leveraged, such as structure in the input or multiple similar forward passes, and show when they fail. 
    Our proposed method utilizes a fast attention-aware recursive clustering method, and compensation of excluded clusters through their mean. The clustering method gives power-of-two cluster sizes, allowing block-sparse attention to match dense attention in GPU throughput. 
    On TabPFN-3, a model where none of the assumptions hold, ClusterAttention is to our knowledge the first method to provide a substantial speedup over the default attention, while consistently keeping over 99\% of its accuracy. On the largest dataset from the TALENT benchmark suite, it makes processing of the training dataset close to 8x faster at nearly 11x attention speedup. ClusterAttention is also competitive with domain-specific methods, while avoiding any of the domain-specific engineering. On video-generation with Wan 2.1-T2V-14B it produces output closer to dense attention at a larger speedup (1.8x vs 1.4x) than SVOO, a leading method in this domain, with both evaluated without offline calibration.
    \footnote{
        Evaluation is currently limited due to time and budget constraints. The code is available on \url{https://github.com/SpoketKasper/ClusterAttention}, where the full preprint history can be found.
    }
\end{abstract}

\newpage
\section{Introduction}
\label{motivation}
Bidirectional attention is a commonly occurring operation in modern transformer-based models, where no causal structure is necessary. Examples of this are models for tabular data \cite{grinsztajn2026tabpfn3technicalreport}, vision transformers (for example applied to pathology \cite{Xu2024}), video generation models \cite{hu2025ultragenhighresolutionvideogeneration}, many text embedding models, and genomic models \cite{nguyen2023hyenadnalongrangegenomicsequence}. However, a drawback of this operation is that its computational cost is quadratic in the number of tokens, as each token attends to all tokens (including itself). In all of the mentioned cases, large token counts can occur, which forces methods working in these domains to set limits to the token count, accept the high latency, or fundamentally alter the attention computation.\\

Often, however, the computation can be approximated well at a lower cost. Contemporary methods tend to do this by grouping tokens, performing some operations on the group-level, and performing computation on a token level only sparsely \cite{zhang2025spargeattentionaccuratetrainingfreesparse, luo2026attentionsparsityinputstabletrainingfree, li2026solattnacceleratingvideogeneration, zhou2026svgearparameterfreelinearcompensation}. Some apply a \textbf{sparsity assumption}, that many of the attention weights are small and can be ignored. These methods are often called sparse attention methods \cite{zhang2025spargeattentionaccuratetrainingfreesparse, luo2026attentionsparsityinputstabletrainingfree, yang2026sparsevideogen2acceleratevideo, tan2026adaclusteradaptivequerykeyclustering}. We show that when this assumption holds, tight clusters that allow capturing the attention-mass give a small error at a low computational cost. 
%\ad{Maybe write where we talk about it?}
However, when it does not hold, tight clusters can give a much larger error than random clusters. To counteract this, excluded groups can be compensated for on a group-level \cite{li2026solattnacceleratingvideogeneration, zhou2026svgearparameterfreelinearcompensation, li2026pisapiecewisesparseattention}. We show that when this is used, tighter clusters give a smaller error also for models where the sparsity assumption do not hold, like TabPFN-3. Another assumption we call the \textbf{context assumption}, and refers to assumptions about the context the attention layer is used in. We find that contemporary methods either assume there is structure in the input, such as order in language or spatial proximity in images \cite{zhang2025spargeattentionaccuratetrainingfreesparse, li2026pisapiecewisesparseattention, li2026solattnacceleratingvideogeneration}, or the possibility to amortize groups formed through slow clustering over repeated forward passes \cite{luo2026attentionsparsityinputstabletrainingfree, tan2026adaclusteradaptivequerykeyclustering, yang2026sparsevideogen2acceleratevideo, zhou2026svgearparameterfreelinearcompensation}. This makes it necessary to perform domain-specific engineering to use the methods, and excludes domains that lack either of these properties, such as tabular data. While previous work has avoided these assumptions \cite{xiong2021nystromformernystrombasedalgorithmapproximating, chen2021scatterbrain, han2023hyperattentionlongcontextattentionnearlinear}, we find that these methods fall short of providing a meaningful speedup while retaining near-dense quality. The solution we present in this work is a fast recursive clustering algorithm based on principal direction divisive partitioning \cite{Boley1998}, which together with attention aware clustering distances can produce useful clusters. The size of these can also be controlled, allowing us to set them to a power-of-2 size and achieve the same GPU throughput as dense attention.\\

On the tabular foundation model TabPFN-3, which satisfies neither of the two assumptions, we find that ClusterAttention is the only method to provide a meaningful speedup over default attention, while consistently retaining at least 99\% of accuracy. On the largest classification dataset in the TALENT \cite{liu2024talenttabularanalyticslearning} dataset suite, ClusterAttention achieves an 8x speedup of training-dataset processing, with the attention itself being 11x faster. We also see that our method is competitive with leading domain-specific methods where assumptions do hold. On the video-diffusion transformer Wan 2.1-T2V-14B, ClusterAttention achieves output closer to dense attention than SVOO \cite{luo2026attentionsparsityinputstabletrainingfree}, both without offline calibration, while providing a meaningfully higher speedup at 1.8x over 1.4x. The main shortcoming of ClusterAttention compared to those leveraging structure is that clustering overhead makes it slow at low tokens counts, so it only starts providing a speedup over fast quantized dense attention somewhere above 20,000 tokens. Lowering this overhead is a possible direction for future work.

\section{Related work}
\label{related-works}
FlashAttention \cite{dao2022flashattentionfastmemoryefficientexact, dao2023flashattention2fasterattentionbetter, shah2024flashattention3fastaccurateattention, zadouri2026flashattention4algorithmkernelpipelining} is an impactful line of work that speeds up exact attention by carefully considering how the operation happens on a GPU. SageAttention \cite{zhang2025sageattentionaccurate8bitattention, zhang2025sageattention2efficientattentionthorough, zhang2026sageattention3microscalingfp4attention} is a parallel line focused on the quantization axis, also leveraged by FlashAttention. These methods are often orthogonal to other speedup methods, as demonstrated by SpargeAttention \cite{zhang2025spargeattentionaccuratetrainingfreesparse}, from which we build our block-sparse kernel.\\

Trainable methods for attention speedup can be aggressive, as they allow the model weights to adapt during training. This can be through, for example, low-rank attention approximation \cite{wang2020linformerselfattentionlinearcomplexity} or sparsity \cite{lu2025mobamixtureblockattention}. 
A well known sparse example is DeepSeek-V3.2 \cite{deepseekai2025deepseekv32pushingfrontieropen}, though later models in the same line combine compressed sparse attention layers with heavily compressed ones, allowing for non-sparse attention as well \cite{deepseekai2026deepseekv4highlyefficientmilliontoken}. Within video-generation, works include VSA \cite{zhang2025vsafastervideodiffusion} and BSA \cite{zhan2025bidirectionalsparseattentionfaster}, which achieve large speedups. The drawback of these methods is that the training involved can make using them a significant engineering or research effort, limiting their adoption.\\

Training-free methods often rely on input structure to group tokens \cite{li2026solattnacceleratingvideogeneration, zhang2025spargeattentionaccuratetrainingfreesparse, jiang2024minference10acceleratingprefilling, xu2025xattentionblocksparseattention, li2026pisapiecewisesparseattention}. Some methods targeting video-generation cluster keys and queries separately, assign a subset of key-clusters to each query-cluster, and then amortize the clustering over several diffusion steps \cite{yang2026sparsevideogen2acceleratevideo, tan2026adaclusteradaptivequerykeyclustering, luo2026attentionsparsityinputstabletrainingfree}. Several methods from both of these lines compensate for excluded groups on a group-level \cite{zhou2026svgearparameterfreelinearcompensation, li2026solattnacceleratingvideogeneration, li2026pisapiecewisesparseattention}. There are also methods that are training-free while requiring neither structure nor amortization \cite{xiong2021nystromformernystrombasedalgorithmapproximating, han2023hyperattentionlongcontextattentionnearlinear, vyas2020fasttransformersclusteredattention, chen2021scatterbrain, daras2020smyrfefficientattentionusing}. Nyströmformer \cite{xiong2021nystromformernystrombasedalgorithmapproximating} uses Nyström theory to estimate the attention matrix from a set of anchor-points, while HyperAttention \cite{han2023hyperattentionlongcontextattentionnearlinear} jointly clusters keys and queries, performs exact attention within each joint cluster, and then estimates the tail of the attention interactions with random samples. In our testing (Appendix \ref{same-niche-methods}), these are not able to provide a meaningful speedup while retaining near-dense quality. Our method is most similar to SVG-EAR \cite{zhou2026svgearparameterfreelinearcompensation}, but uses a different and faster clustering method based on principal direction divisive partitioning \cite{Boley1998}, which allows it to speed up attention at a high quality without amortization of the clustering. The clustering produces clusters of preselected sizes, and choosing a hardware-friendly power of two lets the block-sparse attention run at dense-attention throughput. Similarly to other works \cite{daras2020smyrfefficientattentionusing, han2023hyperattentionlongcontextattentionnearlinear, luo2026attentionsparsityinputstabletrainingfree}, we use cluster-distances that take into account the roles of keys and queries in attention as described in Subsection \ref{target-space-clustering}.

\section{Error analysis of sparse attention variations}
\label{sparse-errors}
In this section we analyze the approximation error of sparse attention with and without compensation of excluded clusters through their centroid, motivating why we use the compensation. We start by analyzing the purely sparse attention. 
Defining $v_i$ for the value of token $i$ and the weight it receives from a query we are inspecting as $w_i$, we can form this query's attention output as 
\begin{equation}
    o = \sum_i w_i v_i,
\end{equation}
where $\sum_i w_i=1$. 
If we select a subset of key-values $S$ to attend to, the output from this attention is 
\begin{equation}
    o_S = \frac{\sum_{i\in S} w_i v_i}{\sum_{i\in S}w_i} = \frac{\sum_{i\in S} w_i v_i}{w_S},
\end{equation}
where we define $w_S:=\sum_{i\in S} w_i$. Defining the set of excluded tokens as $\bar{S}$ and the output of attention on only those as $o_{\bar{S}}$ we then see that
\begin{equation}
    o = w_S o_S + (1 - w_S)o_{\bar{S}}.
\end{equation}
Thus, we find that the error is 
\begin{equation}
    o - o_S = (w_S-1) o_S + (1-w_S)o_{\bar{S}} = (1-w_S) (o_{\bar{S}}-o_S),
\end{equation}
i.e. two factors control the attention error. The first is the missed attention-mass, which is a common focus in sparse attention methods. The second is the deviation between attention-weighted mean values in the included and excluded sets. 
We note that in the case where key-value covariance is high, sparse attention that selects the keys with the highest dot-product with the query induces a large deviation in the second term. In this work, we encounter the counterintuitive result that randomly assigned clusters, for which centroid-based selection between query- and key-clusters is carried out, can outperform sophisticated clustering methods targeting a high attention-mass recall for a given set of keys, i.e. a small left factor in the error expression, as the random clusters naturally induce a small second factor. This was seen both in the vision-transformer DINOv2 \cite{oquab2024dinov2learningrobustvisual}, and in the tabular data-transformer TabPFN-3 \cite{grinsztajn2026tabpfn3technicalreport}, as seen in Subsections~\ref{dino-results} and \ref{tabpfn-results}. We note that while the first factor is intuitively decreased by tight query- and key-clusters, addressing the second factor is less straight-forward.
\\

We now derive the error with compensation, and show that in this case the error reliably shrinks with tighter clusters. The output with compensation is 
\begin{equation}
    \hat{o}_S = \frac{\sum_{i\in S} w_i v_i + \sum_{j\in \bar{S}}w_{c(j)}\bar{v}_{c(j)}}{\sum_{i\in S}w_i + \sum_{j\in \bar{S}}w_{c(j)}},
    \label{ohat-eq}
\end{equation}
where we can see that token $j$ in an unselected cluster is represented through its key-centroid as $w_{c(j)}$ and value-centroid $\bar{v}_{c(j)}$. Note that $w_{c(j)}$ means the "weight" of the centroid if normalization were the same as for dense attention. 
In this work, we will refer to this way of including unselected clusters as \textit{striped mean-compensation} (SMC), as we compensate for excluding the tokens by including them through their mean, producing a visually striped attention matrix. 
We can derive (as done in Appendix~\ref{err-mean-comp}) that the error for this is exactly 
\begin{equation}
    o - \hat{o}_S = \frac{1}{\hat{d}_S}
    \sum_{c\in C_{\bar{S}}} |c|(\delta_c(\bar{v}_c-o) + \mathrm{Cov}_c(w, v)),
\end{equation}
%where $\hat{o}_S$ is the output of the sparse attention with compensation from the centroids, 
where $C_{\bar{S}}$ is the set of clusters excluded from the sparse attention, $|c|\delta_c$ is the attention-mass error of a cluster caused by Jensen's inequality in the softmax exponential, and $\hat{d}_S$ is the denominator in Equation~\eqref{ohat-eq}. 
With most of the attention-mass covered densely and reasonably tight clusters, the underestimation in $\hat{d}_S$ should be small, something we also see empirically. Approximating this with 1, we can simplify to 
\begin{equation}
    o - \hat{o}_S \approx
    \sum_{c\in C_{\bar{S}}} |c|(\delta_c(\bar{v}_c-o) + \mathrm{Cov}_c(w, v)).
\end{equation}
We call the first term the \textit{Jensen} term and the second the \textit{covariance} term. Clearly, the second factor in the Jensen term is not easily modified, except maybe by making extremely loose clusters to move the centroids towards the value mean. However, the first factor is clearly minimized by clusters that are tight in the keys. The covariance term scales with the spread of both keys and values, so it is minimized by clusters that are tight in both keys and values. Empirically on DINOv2, it appears that the covariance term is larger than the Jensen term by roughly an order of magnitude, although the testing of this has not been extensive. This was measured by artificially removing the component through a dense computation, and inspecting the output error. 

\section{Method}
In this section we describe ClusterAttention in detail. On a high level, it performs the following steps: 1. Cluster keys and queries separately. 2. Assign to each query-cluster a subset of the key-clusters and let the queries attend to these keys with per-token resolution, which is known as \textit{sparse} attention. 
3. Approximate the interactions between the queries and the unassigned clusters on a key-cluster level. 
We note here that everything is performed at a per-attention head resolution. 
The sparse attention is carried out using a fork of the quantized block-sparse attention kernel from SpargeAttn \cite{zhang2025spargeattentionaccuratetrainingfreesparse}, where our modifications include an option to return the attention denominator, and the possibility to apply a mask to the keys. Since this kernel only allows the cluster sizes of 128 for keys and 64 for queries, we use these in our evaluation, though we note that multiples of these cluster sizes can quite trivially be used.

\subsection{Clustering}
\label{recursive_splitting}

The clustering is based on a recursive splitting method, which can be performed with non-Euclidean distances that take into account the downstream usage. We describe several variations.

\subsubsection{Recursive splitting}

\paragraph{Basic recursive splitting}
The splitting is a variation on principal direction divisive partitioning (PDDP) \cite{Boley1998}, applied to keys and queries separately. 
With \textit{split} we mean partitioning a set of vectors $\mathcal{S}$ into two using a hyperplane. 
We set the normal vector of the hyperplane to the first principal component of $\mathcal{S}$, estimated from a subsample of vectors using power-iteration. The subsample is of size $\min \{ad, |\mathcal{S}|\}$, $d$ being the dimension of the vectors, and $a$ being set to 1. $\mathcal{S}$ is then projected onto the found direction, and the split is placed so the multiple of $c$ closest to $|\mathcal{S}|/2$ vectors fall below it. 
We find the split value using two 8-bit radix-select passes on the projections. If multiple projections have the exact split value, they are partitioned arbitrarily. 
The splitting is repeated recursively until all sets, or all but one set if the total number of vectors is not divisible by $c$, contain $c$ vectors. 

\paragraph{Diagonalized recursive splitting}

Instead of approximating the first principal component and projecting the vectors onto it at every split, we can choose to split along a coordinate of the vectors, avoiding the power iteration and projection. The coordinate is chosen as the one along which the variance is highest in a subsample of $\min \{d, |\mathcal{S}|\}$ vectors. To make the coordinates good split-directions, we rotate the vectors into the principal component basis of the full set of vectors before starting to partition them. We find the basis through eigendecomposition of the covariance matrix. While this method cannot find the exact highest-variance direction at every split, it gives a more statistically stable estimation problem, and each split becomes faster. 
During the development process, we settled on this approach as it appeared to give better results at a lower latency, though this may need more extensive testing to determine conclusively.

\subsubsection{Attention-aware distances}
\label{target-space-clustering}
Instead of using Euclidean distance in the clustering, we can define distances informed by attention. For clarity of notation, we exclude the $1/\sqrt{d}$ temperature scaling of logits in the following calculations, considering it as part of the softmax operation itself. 

\paragraph{Key clustering}
We note that two keys behave similarly in attention if they get the same logits for all queries, i.e. $k_1 \cdot q \approx k_2 \cdot q,\ \forall\ q \in \mathbf{Q}$, where $\mathbf{Q}$ is the set of queries for an attention head in a given forward pass. If we arrange the $n$ queries into the query matrix $Q$, we see that the average squared difference between the logits of two keys over $\mathbf{Q}$ is
\begin{equation}
    ||Qk_1-Qk_2||^2 / n = ||Q(k_1-k_2)||^2 / n = (k_1-k_2)^T\frac{Q^TQ}{n}(k_1-k_2) := d_q(k_1,k_2)^2,
\end{equation}
where we have defined the pseudo-metric $d_q$ with metric matrix $M = Q^TQ/n$. 
Since queries often span only a subspace of $\mathbf{R}^d$, it is likely that $Q$ does not have full column rank, making $d_q$ not a proper metric. 
This means that distinct keys can have a $d_q$ of 0, which is a useful property as it inhibits clustering along dimensions that do not matter for attention. 
$M$ is positive semi-definite as $x^T(Q^TQ/n)x=||Qx||^2/n\geq0$, so we can factor it as $R_q^TR_q$. While any factorization of this form works, we use eigendecomposition as it reliably handles singular matrices. 
Using this, we see that 
\begin{equation}
    d_q(k_1,k_2)^2 = (k_1-k_2)^TR_q^TR_q(k_1-k_2) = ||R_q(k_1-k_2)||^2 = d(R_qk_1,R_qk_2)^2
\end{equation}
where $d$ is the Euclidean distance. Thus, for diagonalized clustering, we can project through $R_q$ and then proceed normally. Note that $M$ is formed on-the-fly with actual queries, not precalibrated.

\paragraph{Query clustering}
We note that there is an asymmetry between the roles of keys and queries in softmax. The shift invariance over the dot-products means that two queries with identical attention distributions can be arbitrarily far apart in Euclidean distance, if the key distribution has zero-variance directions. The coupling across keys also means that two queries having the same dot-product with a key do not necessarily give it the same attention weight. We derive representations for the query-clustering based on the following arguments: for top-$k$ assignment, we want to cluster queries that have similar rankings of the key-clusters (which we assume the ranking of the keys is a good proxy for), and for adaptive assignment, we want to cluster queries that have similar attention distributions over the keys. We find that both representations end up using the metric-matrix $K_c^TK_c$, where $K_c$ is the key-matrix with all dimensions centered. This centering quotients out the shift-invariance. For diagonalized recursive splitting, the queries are projected through $R_k$, defined in the same way as $R_q$. This produces the representation for the adaptive case, while for the top-$k$ case the projection is followed by normalization. The derivations are found in Appendix~\ref{ranking-objective} and \ref{same-softmax-objective}.

\subsection{Cluster assignment for sparse attention}

We base the selection on dot products between the query-cluster-centroids $\bar{q}$ and key-cluster-centroids $\bar{k}$. 
Due to linearity, the resulting quantity is exactly the average dot product between the vectors in the clusters, so the average logit up to a constant factor. 
However, due to the nonlinearity of softmax, ranking clusters by this quantity is not exactly representative of their relative post-softmax scores. The unnormalized (over keys) attention score between a key and a query (with $\sqrt{d}$-normalization) is $e^{q^\top k/\sqrt{d}}$. Normalization over keys clearly does not change the ranking of scores. Using Jensen's inequality for averages, and noting that the exponential function is strictly convex, we see a key-cluster $\keycluster$ and query-cluster $\querycluster$
\begin{equation}
    \frac{1}{|\querycluster||\keycluster|} \sum_{q \in \querycluster} \sum_{k \in \keycluster} e^{q^\top k/\sqrt{d}}
    \geq \exp\left(\frac{1}{|\querycluster||\keycluster|} \sum_{q \in \querycluster} \sum_{k \in \keycluster} \frac{q^\top k}{\sqrt{d}}\right)
    = e^{\bar{q}^\top \bar{k}/\sqrt{d}}.
\end{equation}
The equality only happens when all pairwise dot products are the same, for example when vectors in each of the clusters are the same. Roughly, it underestimates the attention-mass more for looser clusters. We tried approaches to account for the spread, but these did not give meaningful improvements.\\

We then select the highest-scoring key-clusters for each query-cluster, either the top-$k$ clusters for a given $k$, or until an approximate attention-mass recall threshold is passed based on attention weights over the centroid-scores. We find the score cutoff using a two-pass histogram method with $n_\mathrm{bins}=128$. We use two histograms with shared bin-edges for the cluster-count and attention-mass criteria, so each pass finds one cutoff-bin for each. We refine within the cutoff-bin with lower scores to ensure that both criteria are satisfied, and the final cutoff is the lower edge of the lower cutoff-bin from the second pass. For the top-$k$ assignment, the attention-mass recall threshold is set to 0, and for adaptive assignment, the minimum count is set to 16, i.e. a small non-zero number, as a robustness measure that costs very little in latency.

\subsection{Computational complexities}
Table \ref{complexities-table} summarizes the computational complexities for the different steps, per-head per-layer. We also show the complexity of dense attention for reference. $n$ is the token count, $d$ is the embedding-dimension, and $c_k$ and $c_q$ as the cluster sizes for keys and queries respectively. The clustering is performed for both, so we leave it as $c$. $\bar{f}$ is the average density of the sparse attention over all queries. 
The non-trivial expressions are derived in Appendix~\ref{complexities}. The assignment complexity has one term for scoring the query-centroids against the key-centroids, and one for the partitioning histogram passes. The cost of forming the centroids is small and thus neglected. 
Note that the expression can miss memory transfers without computation, like picking out a coordinate in the diagonalized recursive splitting. We keep asymptotically redundant terms in some expressions, where differing constant factors make them meaningful. 

\begin{table}[h]
    \centering
    \caption{Per-head computational complexity of each step of ClusterAttention, with dense attention for reference. Derivations in the non-trivial cases are found in Appendix~\ref{complexities}.}
    \label{complexities-table}
    \begin{tabular}{ll}
        \toprule
        Step & Complexity \\
        \midrule
        \textbf{Clustering} & \\
        \quad Transformations & $\mathcal{O}(nd^2 + d^3)$ \\
        \quad Diagonalized recursive splitting & $\mathcal{O}(nd^2 + d^3 + d^2\log_2(n/c) + n\log_2(n/c))$ \\
        \textbf{Sparse attention} & \\
        \quad Assignment & $\mathcal{O}\left(d(n/c_k)(n/c_q) + (n/c_k)(n/c_q)\right)$ \\
        \quad Attention computation & $\mathcal{O}(\bar{f}n^2d)$ \\
        \textbf{SMC} & $\mathcal{O}(dn(n/c_k))$ \\
        \midrule
        \textit{Dense attention} & $\mathcal{O}(n^2d)$ \\
        \bottomrule
    \end{tabular}
\end{table}

\section{Results}
ClusterAttention has been tested in three domains: vision-transformers, transformers for tabular data, and diffusion transformers for video generation. The vision transformer, DINOv2-L \cite{oquab2024dinov2learningrobustvisual}, is tested on image resolutions beyond what it was originally trained on. While there is indication that its performance improves with increased resolution outside its native range \cite{bensaid2025novelbenchmarkfewshotsemantic}, it is not clear how far this goes. Therefore, it is considered only for representation distortion instead of downstream tasks. The other models, TabPFN-3~\cite{grinsztajn2026tabpfn3technicalreport} and Wan 2.1-T2V-14B~\cite{wan2025wanopenadvancedlargescale}, are tested in their normal operational range, so we can inspect their downstream performance and draw direct links to how they will operate in practice. As points of comparison, we use SpargeAttn \cite{zhang2025spargeattentionaccuratetrainingfreesparse} and SVOO \cite{luo2026attentionsparsityinputstabletrainingfree}. We also use as baselines the default dense attention of each model (cuDNN backend for DINOv2 and TabPFN-3, FlashAttention-2 for Wan 2.1-T2V-14B), dense attention using the fast quantized kernel of SageAttention2 \cite{zhang2025sageattention2efficientattentionthorough} (to compare the latency given equal quantization), and a method using random clusters and top-$k$ routing through centroid-scores which was found to perform surprisingly well for reasons discussed in Section~\ref{sparse-errors}. SpargeAttn is only tested on DINOv2, as it does not support the unstructured inputs of TabPFN-3, and is not competitive with SVOO on video-generation in the testing performed by the latter. Descriptions of configuration of SpargeAttn and SVOO, and experimental setups for each domain, are found in Appendix~\ref{experimental-setups}. Tests of the methods targeting the same setting as ClusterAttention, which were found not to be competitive, are found in Appendix~\ref{same-niche-methods}.

\subsection{DINOv2-L}
\label{dino-results}
\begin{figure*}[!h]
  \centering
  \begin{subfigure}[b]{0.32\textwidth}
    \includegraphics[width=\textwidth]{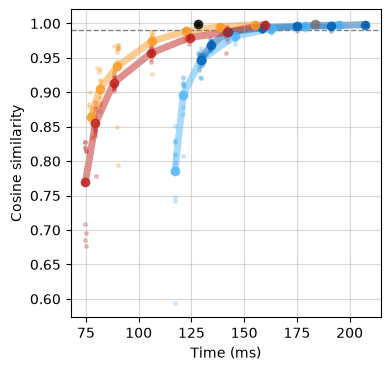}
    \caption{2072 by 2072 pixels.}
    \label{baseline_comparison_small}
  \end{subfigure}
  \hfill
  \begin{subfigure}[b]{0.32\textwidth}
    \includegraphics[width=\textwidth]{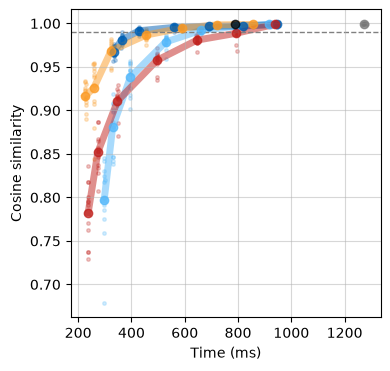}
    \caption{3500 by 3500 pixels.}
    \label{baseline_comparison_medium}
  \end{subfigure}
  \hfill
  \begin{subfigure}[b]{0.32\textwidth}
    \includegraphics[width=\textwidth]{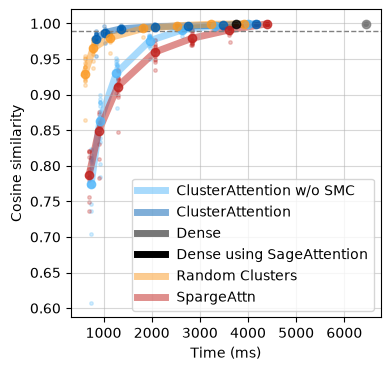}
    \caption{5306 by 5306 pixels.}
    \label{baseline_comparison_large}
  \end{subfigure}
  \caption{Comparison between some ClusterAttention configurations and baselines. Pareto front on latency vs. quality, measured in average cosine-similarity with the output from dense attention.}
  \label{pareto-baselines}
\end{figure*}

We evaluate ClusterAttention on computer vision using DINOv2-L, which is a vision transformer developed by Meta, on images from DIV8K \cite{9021973} with a resolution of at least 5306 pixels on the shorter axis. All evaluations are done as Pareto-fronts of representation distortion versus latency for the full forward pass. 
The distortion is measured as embedding cosine similarity with dense, averaged over all the tokens/patches. The results are seen in Figure \ref{pareto-baselines}. To be able to quantify the actual usefulness of the methods, we have drawn a dashed line marking 0.99 cosine similarity with dense attention, a reference operating point for comparing latencies. This indicates which part of the Pareto-front is the most interesting. 
This threshold is just a suggestion. The actual appropriate threshold depends on the downstream application and how sensitive it is to representation distortion, and needs to be measured empirically from case to case. The adaptive methods, including SVOO, are shown in Appendix \ref{svoo-dino} to keep Figure \ref{pareto-baselines} legible. These were not competitive. 

\subsection{TabPFN-3}
\label{tabpfn-results}

TabPFN-3 is the latest in the series of tabular foundation models from Prior Labs. We tested applying ClusterAttention to the in-context-learning attention layers of this model, in the context of the 6 largest classification datasets from the TALENT benchmark \cite{liu2024talenttabularanalyticslearning}. We draw Pareto-fronts of accuracy normalized by that of dense attention, versus latency of the train-to-train processing in the model backbone. We used a single forward pass to generate each prediction, so no pre-processing ensembling. The results for some configurations are shown in Table~\ref{tabpfn-table}, with full Pareto-sweeps in Appendix~\ref{tabpfn-pareto-appendix}. We have bolded speedups over dense, and accuracies over a suggested operating point of 0.99 of the accuracy achieved with dense attention. On the largest dataset, we measure the latency of all parts except for the attention backend to be roughly 2.1 seconds out of the total 51.8 seconds.

\begin{table}[t]
    \centering
    \caption{TabPFN-3 on the 6 largest classification datasets from the TALENT suite. Dense shows absolute latency in seconds and accuracy in percent. All other entries show speedup/accuracy, both relative to dense. Speedups above 1 and relative accuracies of at least 0.99 are bold. CA and Sage2 are short for ClusterAttention and SageAttention2 respectively. The percentage after @ is the top-$k$ percentage for all methods except for SVOO where it is the attention-mass recall. We show the configuration of each method that crosses 0.99 for at least half of the datasets, plus ClusterAttention at 10\% which crosses it for all.}
    \label{tabpfn-table}
    \small
    \begin{tabular}{r|ccccc|c}
        \toprule
        $n$ & CA@5\% & CA@10\% & SVOO@80\% & Random@40\% & Sage2 & Dense \\
        \midrule
        93k & \textbf{3.39}/\textbf{0.998} & \textbf{3.11}/\textbf{1.000} & 0.60/0.936 & \textbf{2.38}/0.969 & \textbf{1.29}/\textbf{1.000} & 1.26/67.4 \\
        96k & \textbf{2.78}/0.983 & \textbf{2.58}/\textbf{0.992} & 0.59/\textbf{0.991} & \textbf{2.09}/\textbf{0.995} & \textbf{1.26}/\textbf{1.000} & 1.29/85.8 \\
        98k & \textbf{3.49}/\textbf{0.994} & \textbf{3.17}/\textbf{0.997} & 0.57/0.918 & \textbf{2.40}/0.975 & \textbf{1.31}/\textbf{1.000} & 1.33/74.7 \\
        162k & \textbf{4.14}/\textbf{0.999} & \textbf{3.67}/\textbf{0.999} & 0.77/\textbf{0.991} & \textbf{2.43}/\textbf{0.996} & \textbf{1.29}/\textbf{1.000} & 3.62/86.8 \\
        430k & \textbf{9.38}/\textbf{0.996} & \textbf{7.14}/\textbf{0.997} & \textbf{1.42}/\textbf{0.997} & \textbf{3.06}/\textbf{0.997} & \textbf{1.36}/\textbf{1.000} & 23.6/96.8 \\
        634k & \textbf{10.44}/\textbf{0.994} & \textbf{7.76}/\textbf{0.998} & \textbf{1.44}/\textbf{0.995} & \textbf{3.16}/\textbf{0.999} & \textbf{1.33}/\textbf{1.000} & 51.8/74.1 \\
        \bottomrule
    \end{tabular}
\end{table}

\subsection{Wan 2.1-T2V-14B}
We evaluate ClusterAttention on 720p video generation with Wan 2.1-T2V-14B, a video generation diffusion-transformer developed by Alibaba. 
The prompts are from a list of 12 prompts from OpenSora 1.0 used by SpargeAttn among others, which we include in Appendix \ref{prompts}. We evaluate on 
the first 5 prompts from this list. ClusterAttention is set to attend to 20\% of clusters. We apply it as-is, without any cluster caching or other stateful techniques that are common for sparse video generation methods. For example, SVOO recomputes clusters only after every 20th diffusion step. We use the metrics peak signal-to-noise ratio (PSNR), learned perceptual image patch similarity (LPIPS) \cite{zhang2018unreasonableeffectivenessdeepfeatures}, and
Structural Similarity Index Measure (SSIM) \cite{1284395}, to measure the deviation between the output of a dense generation and approximate ones. The results are shown in Table \ref{dit}. The speedups for SVOO are lower than those reported by Luo et al., which appears to be an effect of dense attention being markedly faster in our testing, as the latency for SVOO is similar in our measurements to theirs. 
\begin{table}[h]
    \centering
    \caption{Comparison of video quality and generation latency on 5 prompts. Left number for every metric is ClusterAttention, right is SVOO.}
    \label{dit}
    \begin{tabular}{c|c|c|c|c}
        Prompt & PSNR$\uparrow$ & SSIM$\uparrow$ & LPIPS$\downarrow$ & Speedup \\
        \hline
        1 (night city) & \textbf{20.4}/19.4 & \textbf{0.764}/0.743 & \textbf{0.183}/0.206 & \textbf{1.81}/1.43\\
        2 (toy boat) & \textbf{24.5}/24.2 & \textbf{0.906}/0.899 & \textbf{0.0562}/0.0658 & \textbf{1.80}/1.42\\
        3 (waterfall) & \textbf{27.0}/26.8 & 0.870/\textbf{0.883} & \textbf{0.0671}/0.0814 & \textbf{1.87}/1.42\\
        4 (milky way) & \textbf{30.0}/28.3& \textbf{0.918}/0.897 & 
        \textbf{0.0634}/0.0716 & \textbf{1.83}/1.46\\
        5 (turtle) & \textbf{21.7}/21.4& \textbf{0.706}/0.691 & \textbf{0.120}/0.139 & \textbf{1.80}/1.44
    \end{tabular}
\end{table}

\subsection{Ablations}
We compare eight versions of ClusterAttention, one for each combination of the following three choices: transforming keys and queries before clustering or not, top-$k$ or adaptive selection, and SMC or not. The evaluation is performed on DINOv2 with the setup from Subsection \ref{dino-results}. 
ClusterAttention configured to use transforms, top-$k$ selection, and SMC is generally Pareto-dominant. The transforms improve results for almost all setups. The exceptions are on the smallest image size at very high sparsity, where their overhead dominates, and with SMC applied at the largest image size at low sparsities. SMC improves quality at all settings at a small latency cost. Without SMC, the selection matters little, but with it, top-$k$ performs much better than adaptive, which we discuss in Appendix~\ref{compensation-ablation}. 
Several variations are also included for TabPFN-3 in Figure \ref{accelerometer} and agree with these conclusions, indicating that they generalize beyond DINOv2. However, models with very different attention patterns may give different results. Pareto sweeps for the ablations are found in Appendices~\ref{transform-ablations}, \ref{compensation-ablation} and \ref{selection-ablation}.
\\

We also ablate the method with random clusters on DINOv2. We find that it performs the best when both keys and queries use random clusters, and when using top-$k$ assignment over random or adaptive assignment. Thus, the shown baseline is the best configuration. It performs worse if SMC is applied, which is consistent with the error analysis in Section~\ref{sparse-errors}.

\subsection{Latency scaling}
\begin{figure}[h]
  \centering
  \begin{subfigure}[b]{0.46\textwidth}
    \includegraphics[width=\textwidth]{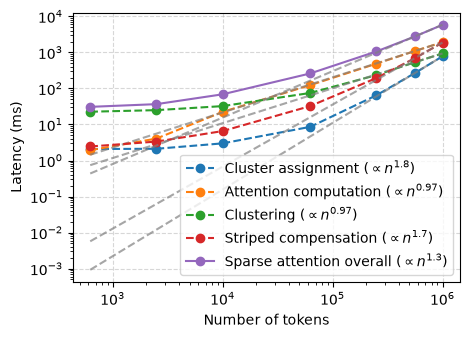}
    \caption{$k$ fixed to 64 blocks of 128 keys.}
    \label{latency-fixed}
  \end{subfigure}
  \hfill
  \begin{subfigure}[b]{0.46\textwidth}
    \includegraphics[width=\textwidth]{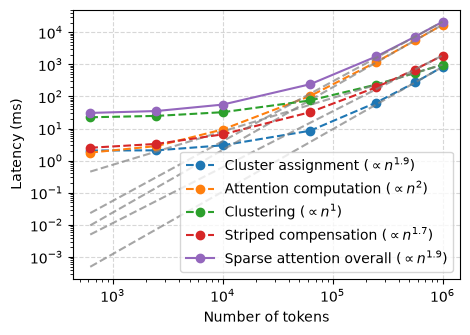}
    \caption{$k$ set so the model attends to 10\% of all keys.}
    \label{latency-10perc}
  \end{subfigure}
  \caption{Latency scaling for ClusterAttention. Timing averages over 3 runs on the first image from DIV8K after shuffling.}
  \label{scaling}
\end{figure}

Figures \ref{latency-fixed} and \ref{latency-10perc} 
show two cases for how the latency ClusterAttention scales with the number of tokens, with a breakdown over its parts, as well as a fitted line in log-log-space between the last two measurements. In Figure \ref{latency-fixed}, 
the number of tokens attended to was fixed to 64 clusters of 128 tokens, so 8192 tokens. This simulates deployment on a model with sparse attention patterns, or that in general works well with highly sparse attention. An example of such a model is DeepSeek-V3.2 \cite{deepseekai2025deepseekv32pushingfrontieropen}, which attends to a fixed 2048 tokens shared over all heads in each layer. This setting is both where sparse or approximate attention can give the largest speedup, and the most adversarial case to overhead. In Figure \ref{latency-10perc} 
we fix that ClusterAttention attends to 10\% of all keys. This simulates deployment on a model where attention is less sparse, such as DINOv2 and TabPFN-3.

\section{Discussion}
\label{discussion}
\subsection{Overall performance}

ClusterAttention is Pareto-dominant in most of the tests involving larger token counts. 
Figure~\ref{baseline_comparison_small} suggests that the crossover at the operating point versus SpargeAttn lies just above the small image size, though the exact value would depend a lot on the chosen operating point, increasing with a lower quality requirement. 
On TabPFN-3, ClusterAttention attending to 10\% of clusters consistently retains over 99\% relative accuracy, with a relative speedup of the train-to-train backbone that grows from around 2.5x to almost 8x with dataset size. 
On the largest dataset, the attention itself is sped up almost 11x compared to default dense attention, or 8x compared to SageAttention2.
On Wan 2.1-T2V-14B, ClusterAttention provides overall better quality retention than SVOO without offline profiling, while also providing a higher speedup. Due to its large clustering overhead, SVOO is not competitive in the other domains, as they fall into the single-pass setting which it was not developed for.
%\kn{notera att clusterattention är stabil mellan runs på samma bild?}

\subsection{Latency in detail}
\label{latency-dicussion}
Figure~\ref{scaling} shows that clustering overhead dominates at low token counts for both deployment examples. The few-token clustering overhead appears to be dominated by the eigenvalue decompositions. 
It is approximately flat up to around 10,000 tokens, where it starts curving upward, and between 562,500 and 1 million tokens grows slightly below linearly. 
In the case with a fixed low $k$, the sparse attention computation overtakes clustering as the largest latency component at around 20,000 tokens. At 1 million tokens, the quadratic SMC catches up, and would dominate above this. With highly sparse attention, applying the compensation more selectively may be the most impactful improvement at large token counts. 
In the case with attention to 10\% of tokens, the sparse attention overtakes clustering slightly later, but then dominates. 
Cluster assignment grows at a rate approaching quadratic, but due to its small constant it is essentially the smallest component across all token counts.

\section{Future work}
\label{future-work}
\subsection{Improvements}
We identify four main axes of improvements for ClusterAttention: faster clustering, better clustering, better compensation of excluded clusters, and better assignment. The latency observations show that in the lower token-count regime (roughly below a few tens of thousands), clustering makes up the majority of the latency. While we have experimented with faster eigenvalue decomposition kernels that lower the overhead by roughly 10 ms, providing a meaningful speedup over SageAttention2 at low token counts would require a much more streamlined implementation of ClusterAttention overall.\\

In the higher token-count regime (up to roughly a million tokens), latency is dominated by the sparse attention. To lower the latency here, the most impactful changes are those that allow attention to fewer keys at maintained approximation quality. Improvements on any of the three latter axes can do this. On cluster quality, we note in Figures~\ref{adaptive-dino-pareto} and~\ref{accelerometer} that SVOO \cite{luo2026attentionsparsityinputstabletrainingfree} achieves markedly lower error for the same approximate attention budget than adaptive ClusterAttention without SMC, for both DINOv2 \cite{oquab2024dinov2learningrobustvisual} and TabPFN-3 \cite{grinsztajn2026tabpfn3technicalreport}. This indicates that it achieves better clusters, which shows that ClusterAttention has headroom in the cluster-quality. One explanation may be that fixed-size clusters are suboptimal, and we note that the recursive splitting framework does support more flexibly assigned cluster-sizes. An idea for improved clustering through nonlinear transformations is discussed in Appendix~\ref{softmax-aware-clustering}. Better clustering should lead to better compensation and assignment. To directly improve compensation, there are options discussed in the literature, for example in MuSe \cite{mitchell2026multipolesemanticattentionfast} and PISA \cite{li2026pisapiecewisesparseattention}. The former removes all first-order terms in cluster-spread through exponential tilting, though incurs memory transfer costs. The latter adds a global (i.e. approximate) first-order correction to the compensation at a minimal latency cost. On assignment, we note that centroid-to-centroid scores with top-$k$ or adaptive assignment may be weak heuristics for which key-clusters are the most important to attend to in the sparse attention. Other options are discussed in, for example, SVG-EAR~\cite{zhou2026svgearparameterfreelinearcompensation} and Sol-Attn~\cite{li2026solattnacceleratingvideogeneration}. The former tries to directly approximate the error arising from mean-compensation of a given cluster, while the latter mainly aims to lower the latency incurred from assignment.

\subsection{Extension beyond MHA}
In this work, we only use models with multi-head attention (MHA). Extensions to grouped-query attention (GQA) or multi-query attention (MQA) are natural, though some ablation and analysis may be required to determine the optimal approach. 
While query-clustering can be performed in the same way as for MHA, key-clustering requires the choice between averaging the second-moment matrices for all associated query-heads, or performing one key-clustering for each associated query-head, or some other option.

\subsection{Generalization in implementation}
The implementation currently only runs on Nvidia H100 and H200 GPUs, and has only been tuned for the former, which may limit the usability of ClusterAttention.

\subsection{Extended evaluation}
To strengthen the evaluation, it can be expanded to include more images, datasets, and prompts, as well as more baselines, method configurations, and models. It would also be interesting to apply ClusterAttention to domains not included in this work. Furthermore, we would like to investigate how the usage of ClusterAttention affects model training.

\section{Acknowledgements}
We would like to thank the team behind SpargeAttn \cite{zhang2025spargeattentionaccuratetrainingfreesparse} for their excellent sparse attention kernel which we use, and the team behind SVOO \cite{luo2026attentionsparsityinputstabletrainingfree} for quickly responding to our questions. 

\clearpage
\newpage

% \bibliographystyle{ieeetr}
% \bibliographystyle{abbrv}
% \bibliography{references}
\printbibliography

\newpage
%\onecolumn
\appendix
\section{Error of mean-compensated sparse attention}
\label{err-mean-comp}
If we partition the excluded attention over a set of clusters each called $c(j)$, and include them through their centroid key and value (with a cluster-size weight-compensation), we find
\begin{equation}
    \hat{o}_S = \frac{\sum_{i\in S} w_i v_i + \sum_{j\in \bar{S}}w_{c(j)}\bar{v}_{c(j)}}{\sum_{i\in S}w_i + \sum_{j\in \bar{S}}w_{c(j)}} = \frac{\sum_{i\in S} w_i v_i + \sum_{j\in \bar{S}}w_{c(j)} \bar{v}_{c(j)}}{w_S + \sum_{j\in \bar{S}}w_{c(j)}}.
\end{equation}
We can collect the terms by cluster and get
\begin{equation}
    \hat{o}_S = \frac{\sum_{i\in S} w_i v_i + \sum_{c\in C_{\bar{S}}}|c|w_{c}\bar{v}_c}{w_S + \sum_{c\in C_{\bar{S}}} |c|w_{c}} = \frac{\hat{n}_S}{\hat{d}_S},
\end{equation}
where we have defined shorthands for the numerator and the denominator. We can do some algebraic tricks with the error expression to find
\begin{equation}
    o - \hat{o}_S = o - \hat{n}_S + \hat{n}_S - \hat{o}_S = (o - \hat{n}_S) + (\hat{d}_S-1)\hat{o}_S.
\end{equation}
Here, we massage the first parenthesis to find
\begin{equation}
    o - \hat{n}_S = 
    \sum_i w_i v_i - (\sum_{i\in S} w_i v_i + \sum_{c\in C_{\bar{S}}}|c|w_{c}\bar{v}_c) = 
    \sum_{i\in \bar{S}} w_i v_i - \sum_{c\in C_{\bar{S}}}|c|w_{c}\bar{v}_c.
\end{equation}
We can collect it all into clusters and write
\begin{equation}
    o - \hat{n}_S =  
    \sum_{c\in C_{\bar{S}}}\sum_{i\in c} (w_i v_i - w_{c}\bar{v}_c).
\end{equation}
We now introduce the average weight of the keys in a cluster $\bar{w}_c$ and with it $\delta_c = \bar{w}_c - w_c \geq0$, where the inequality comes from Jensen's inequality and the convexity of the exponential. We further introduce the perturbations $\epsilon^w_i = w_i - \bar{w}_c = w_i - w_c - \delta_c$, as well as $\epsilon^v_i = v_i - \bar{v}_c$, and rewrite to
\begin{equation}
    o - \hat{n}_S = \sum_{c\in C_{\bar{S}}} \sum_{i \in c} (w_i v_i - w_c \bar{v}_c)= 
    \sum_{c\in C_{\bar{S}}} \sum_{i \in c} (\delta_c\bar{v}_c + \epsilon^w_i\bar{v}_c + \bar{w}_c\epsilon^v_i + \epsilon^w_i\epsilon^v_i)
    ,
\end{equation}
where we get to the right hand side after removing opposite terms. Now, we note that the perturbations are zero-mean, so the two middle terms both sum to the zero-vector, and the expression simplifies to
\begin{equation}
    o - \hat{n}_S = 
    \sum_{c\in C_{\bar{S}}} \sum_{i \in c} (\delta_c\bar{v}_c + \epsilon^w_i\epsilon^v_i) =
    \sum_{c\in C_{\bar{S}}} |c|(\delta_c\bar{v}_c + \mathrm{Cov}_c(w, v)),
\end{equation}
Now, inspecting $\hat{d}_S-1$ we find
\begin{equation}
    \hat{d}_S-1 =  (w_S + \sum_{c\in C_{\bar{S}}} |c|w_{c}) - (w_S + \sum_{c\in C_{\bar{S}}} |c|\bar{w}_{c}) = -\sum_{c\in C_{\bar{S}}} |c|\delta_{c}.
\end{equation}
So, we have 
\begin{equation}
    o - \hat{o}_S = \sum_{c\in C_{\bar{S}}} |c|(\delta_c\bar{v}_c + \mathrm{Cov}_c(w, v)) -\sum_{c\in C_{\bar{S}}} |c|\delta_{c}\hat{o}_S = \sum_{c\in C_{\bar{S}}} |c|(\delta_c\bar{v}_c + \mathrm{Cov}_c(w, v) - \delta_{c}\hat{o}_S),
\end{equation}
which finally comes together to
\begin{equation}
    o - \hat{o}_S = \sum_{c\in C_{\bar{S}}} |c|(\delta_c(\bar{v}_c-\hat{o}_S) + \mathrm{Cov}_c(w, v)).
\end{equation}
This is a nice exact expression, but not very easy to interpret. However, we can make a very similar expression that carries more meaning. If we write the error expression instead as
\begin{equation}
    \hat{d}_S(o - \hat{o}_S) = \hat{d}_So - \hat{n}_S = \hat{d}_So - o + o - \hat{n}_S = (\hat{d}_S - 1)o + (o - \hat{n}_S).
\end{equation}
Based on our computations above, we can substitute in
\begin{equation}
    \hat{d}_S(o - \hat{o}_S) = 
    -\sum_{c\in C_{\bar{S}}} |c|\delta_{c}o + \sum_{c\in C_{\bar{S}}} |c|(\delta_c\bar{v}_c + \mathrm{Cov}_c(w, v)).
\end{equation}
Similarly to above, we can rewrite it
\begin{equation}
    o - \hat{o}_S = \frac{1}{\hat{d}_S}
    \sum_{c\in C_{\bar{S}}} |c|(\delta_c(\bar{v}_c-o) + \mathrm{Cov}_c(w, v)).
\end{equation}

\section{Linear query-representations}
\subsection{Query-representation for adaptive assignment}
\label{same-softmax-objective}
For adaptive assignment, we want to represent queries in a way such that small distances in the representation correspond to small differences in attention weights, the \textit{forward} property, as this means that queries we cluster together will actually produce similar attention distributions. We also want the converse, i.e. that small differences in attention weight correspond to small distances in representation, the \textit{backward} property, as this means that queries producing similar attention distributions will be close in representation, so that clustering does not separate queries that behave similarly. More precisely, these properties correspond to Lipschitz bounds between the representation of the queries and their attention weights. We note that since softmax is Lipschitz continuous \cite{gao2018propertiessoftmaxfunctionapplication}, the forward property holds with the Euclidean representation, and for any linear transform of $q$ from which the logits can be recovered. So, ignoring the Lipschitz constant, the forward property is trivial. On the other hand, the backward property is not globally possible for any non-zero linear representation, as attention weights lie on the probability simplex, so their distances are bounded, while representation distances are unbounded, i.e. no $L$ can for all possible $q_1$ and $q_2$ fulfill 
\begin{equation}
    ||r(q_1)-r(q_2)|| \leq L||\mathrm{softmax}(Kq_1)-\mathrm{softmax}(Kq_2)||.
\end{equation}
To give intuition for this, for small logits (suppressed keys), the softmax derivative gets small, so large changes to a representation that only affects these keys have small effects on the attention distribution. However, a local Lipschitz bound in the backward direction is possible. Consider the representation
\begin{equation}
    r(q) = Kq.
\end{equation}
This does not have the local backward property. In fact, queries with the same attention distribution can be arbitrarily far apart, as the shift invariance of softmax gives that
\begin{equation}
    \mathrm{softmax}(Kq)=\mathrm{softmax}(K(q+\delta))\  \text{iff}\ K \delta = c\mathbf{1},\ \text{for some}\ c\in\mathbb{R},
\end{equation}
meaning that $\delta$ lies in a space where the key distribution has zero variance. A special case of this is when $\delta$ is orthogonal to all keys, which can happen when the key matrix is low-rank, a common situation in practice. More generally, the smaller the variance of the key distribution along a direction, the less it matters for attention weights. This suggests using the covariance matrix of the keys as metric matrix, i.e. $\frac{K_c^TK_c}{n}$ where $K_c$ is the key matrix after mean-centering, meaning that row $i$ is $k_i-\bar{k}$. This implies the representation
\begin{equation}
    r(q) = \frac{K_c}{\sqrt{n}}q.
\end{equation}
With the same $\delta$ as above,
\begin{equation}
    K_c(q+\delta) = K_cq+(K-\mathbf{1}\bar{k}^T)\delta = K_cq+c\mathbf{1}-(\bar{k}\cdot\delta)\mathbf{1}=K_cq,
\end{equation}
where $\bar{k}\cdot\delta=c$ due to the linearity of the mean, so the shifts under which softmax is invariant do not change the representation, i.e. the centering quotients out the equivalence class under softmax. 
Since softmax is Lipschitz continuous \cite{gao2018propertiessoftmaxfunctionapplication} and the logits are linear in the representation, small distances in the representation give small differences in the attention weights, giving the forward property. 
\\

As noted before, the backward property of closeness cannot hold globally. However, differently from $q$ or $Kq$, the implication 
\begin{equation}
    \mathrm{softmax}(K_cq_1)=\mathrm{softmax}(K_cq_2) \implies K_cq_1=K_cq_2
\end{equation}
is valid, since the left-hand side requires that $K_cq_2=K_cq_1+c\mathbf{1}$, but the entries of $K_cq$ sum to 0 for any query as $\sum_i (k_i -\bar{k}) \cdot q = (n\bar{k}-n\bar{k}) \cdot q = 0$, so $c$ must be 0. We expect the lack of a global Lipschitz bound to not undermine the usefulness of the metric; in well-behaved cases and for most queries, where logits remain bounded, no key is fully suppressed, so the bound should not get too loose. 
%But the derivative of the softmax function with respect to very small logits (suppressed keys) is small, meaning that two queries can be far apart along a direction that only affects suppressed keys, without the attention weights changing meaningfully. This is a property of remaining in logit-space and using a linear metric. 

\subsection{Query-representation for key-cluster ranking}
\label{ranking-objective}
A good representation of a query is the relative ranking of the keys by their dot-product with the query. Formally, the representation is
\begin{equation}
    r_{(a,b)}(q) = \text{sign}((K_a-K_b)\cdot q),
\end{equation}
where single indexing into the key-matrix gives a single key, and the representation has one value for each pair from the space of ordered pairs of key-indices with size $n^2$. From here, we relax to the actual differences to get a linear representation, so
\begin{equation}
    r_{(a,b)}(q) = (K_a-K_b)\cdot q,
\end{equation}
noting that we lose scale invariance of $q$. 
At the same time, we gain information on which pairs have larger differences in the dot product, and thus matter more to rank correctly. Incorrect ranking of only pairs with a small difference in the dot product should in the worst case lead to including the wrong clusters around position $k$ in the ranking, which is less impactful than missing top-scoring clusters. 
After the relaxation we still have offset invariance over the keys, which is desirable for comparing rankings. We can note that this representation lives in $\mathbb{R}^{n^2}$, but it is found through a linear transformation, so it can have at most rank $d$. Precisely, the linear transformation is
\begin{equation}
    D_{(a,b),m} = K_{a, m} - K_{b, m}.
\end{equation}
where we impose some ordering over the pairs. Since all geometry in transformed space involves forming $D^TD$, we can study this matrix to find a non-redundant representation. We find
\begin{multline}
    (D^TD)_{jk} = \sum_l D_{lj}D_{lk} = \sum_{(a,b)} (K_{a,j}-K_{b,j})(K_{a,k}-K_{b,k}) = \\
    \sum_{(a,b)} (K_{a,j}K_{a,k} - K_{a,j}K_{b,k} - K_{b,j}K_{a,k} + K_{b,j}K_{b,k}) = \\
    n\sum_{a} K_{a,j}K_{a,k} - \left(\sum_{a} K_{a,j}\right) \left(\sum_{b} K_{b,k}\right) - \left(\sum_{b} K_{b,j}\right) \left(\sum_{a} K_{a,k}\right) + n\sum_{b} K_{b,j}K_{b,k}.
\end{multline}
We then note that 
\begin{equation}
    \sum_{a} K_{a,j}K_{a,k} = (K^TK)_{jk}
\end{equation}
and similar for the sum over $K_{b,j}K_{b,k}$, while 
\begin{equation}
    \left(\sum_{a} K_{a,j}\right) \left(\sum_{b} K_{b,k}\right) = (n\bar{K}_{,j})(n\bar{K}_{,k}) = n^2\bar{K}_{,j}\bar{K}_{,k},
\end{equation}
and similar for the other factored sum, where indexing with $,j$ indicates a \textit{column} of the key matrix. Putting it together, we find 
\begin{equation}
    (D^TD)_{jk} = 2n(K^TK)_{jk} - 2n^2\bar{K}_{,j}\bar{K}_{,k} = 2n(K_c^TK_c)_{jk},
\end{equation}
which is the $2n$ times $j,k$th element of the covariance matrix over the keys. So, using this gives the same geometry up to a scalar factor. Finally, to recover the scaling invariance over $q$ we had in the original ranking metric, we note that scaling $q$ directly scales the representation. So we can simply normalize the representation. With $R_k$ such that $R_k^TR_k = K_c^TK_c$, we get our key-aware query representations as 
\begin{equation}
    r(q) = \frac{R_k q}{|R_k q|}.
\end{equation}

\section{Softmax-aware clustering}
\label{softmax-aware-clustering}
We can rephrase the query-space clustering argument as follows. We consider a representation of $k_1$ that is query-aware as 

\begin{equation}
    r(k_1) = \frac{Q}{\sqrt{n}}k_1,
\end{equation}

i.e., $k_1$ is represented as its dot with all the queries (the logits). This representation lies in $\mathbb{R}^n$, but being a projection from $\mathbb{R}^d$ it occupies a subspace of at most dimensionality $d$. In fact, as attention keys often only occupy a subspace of lower dimensionality than their intrinsic dimensionality, the representation likely occupies a subspace of lower dimensionality than $d$. We can see that we recover the same (pseudo-)metric $M$ as before in this space, since

\begin{multline}
    ||r(k_1)-r(k_2)||^2 = (r(k_1)-r(k_2))^T(r(k_1)-r(k_2)) = \\
    \left(\frac{Q}{\sqrt{n}}k_1 - \frac{Q}{\sqrt{n}}k_2\right)^T\left(\frac{Q}{\sqrt{n}}k_1 - \frac{Q}{\sqrt{n}}k_2\right) = (k_1 - k_2)^T\frac{Q^TQ}{n}(k_1 - k_2).
\end{multline}

To avoid the representation becoming low-rank, we can note that the logits will be passed through a softmax function as part of the attention computation. To mirror this, we apply 
the softmax transform over the keys for each query. This is similar to the kernel-trick in support-vector machines, but with the additional benefit of directly mirroring the downstream usage of the clusters. The specific nonlinearity of softmax means that for a given query, the difference between two larger logits carries more meaning, than the same difference between two smaller logits. Thus, clustering in a \textit{softmax aware} space, by applying a softmax operation over the keys (whether clustering keys or queries), emphasizes tight clustering of keys in regions of the space where they significantly affect the attention values. It can also increase the number of meaningful directions/coordinates to cluster along.\\

Of course, performing this projection is essentially calculating full attention, which is what we want to avoid. To avoid this cost, we can choose a subset of queries as landmarks. This can be done, for example, by random subsampling. We can also project this new representation onto its principal components. There is a risk that the strong selectivity of softmax in this subsampled space could collapse the representations of keys. To mitigate this, the softmax temperature may be increased.

\section{Computational complexities}
\label{complexities}
\subsection{Basic recursive splitting}
\label{cplx-rs}
The number of operations in the dominant parts of split $i$ is proportional to 
\begin{equation}
    2^i \left( \underbrace{pd^2}_{\text{power iteration}} + \underbrace{\frac{n}{2^i}d}_{\text{projection}} + \underbrace{\frac{n}{2^i}}_{\text{partitioning}} \right) = 2^ipd^2 + nd + n,
\end{equation}
where $p$ is the number of power iterations. This is repeated for $i$ from 0 to $\log_2(n/c)-1$, roughly, since $n/c$ is not always a power of two. We get for the power iteration
\begin{equation}
    \label{log-sum}
    \sum_{i=0}^{\log_2(n/c)-1} 2^i p d^2 = p d^2 \sum_{i=0}^{\log_2(n/c)-1} 2^i = pd^2\left(n/c - 1\right)
\end{equation}
and the algorithmic complexity as
\begin{equation}
    \mathcal{O}\left(pd^2\frac{n}{c} + nd\log_2(n/c) + n\log_2(n/c)\right).
\end{equation}
As can be seen, the splitting is dominated by the $n\log_2n$ term in the limit of $n$. In practice, all terms can be meaningful, due to the roughly linear scaling but different constants.

\subsection{Diagonalized recursive splitting}
\label{cplx-vs}

The diagonalization to PC axes requires forming a covariance matrix, eigenvalue-decomposing it, and rotating the vectors into this space. These steps have complexities $\mathcal{O}(nd^2)$, $\mathcal{O}(d^3)$, and $\mathcal{O}(nd^2)$ respectively. The number of operations at split $i$ is then proportional to
\begin{equation}
    2^i \left( \underbrace{\frac{d}{2^i}d}_{\text{variance estimation}} + \underbrace{\frac{n}{2^i}}_{\text{partitioning}} \right).
\end{equation}
Here, subsampling to $d$ vectors was assumed. In total, we get a complexity of 
\begin{equation}
    \mathcal{O}(nd^2 + d^3 + d^2\log_2(n/c) + n\log_2(n/c)),
\end{equation}
since the variance estimation and partitioning are repeated $\log_2(n/c)$ times. We note here that there is no projection term within the splitting, however, there is still a memory access cost associated with picking out a certain coordinate. Given that we do not reorder the vectors during the clustering, clusters mostly end up scattered, so this can involve reading isolated scalars. Since memory is read in larger blocks, the memory transfer cost can be quite large (though still at most as large as in the non-diagonalized case). 

%\newpage
\section{Experimental setups}
\label{experimental-setups}
\subsection{Configuration of other methods}

We include both the top-$k$ and the adaptive versions of SpargeAttn \cite{zhang2025spargeattentionaccuratetrainingfreesparse} for comparison, although it can be noted that Zhang et al. recommend using the top-$k$ version. 
We test both methods with the default configuration settings in the reference implementation, but also test their adaptive method with a lower self-cosine-similarity threshold, as the default setting always runs near dense attention. We run it with row-major clustering for images, as an official implementation of the space-filling-curve form was not found. The difference between row-major and space-filling-curve forms also does not seem to be large in the SpargeAttn paper.\\

For SVOO \cite{luo2026attentionsparsityinputstabletrainingfree}, we use the same settings as they use in their testing. This includes a fixed budget of 1024 key-clusters and 256 query-clusters, which was recommended by the authors, as they found small correlation between optimal cluster counts and token counts in their testing. With that said, we note that we test it outside of its target-application of video generation, and over a larger range of token counts, so this may not be the optimal configuration. We use SVOO without offline profiling, to compare the sparse-attention techniques in isolation. We note that offline sparsity profiling of a model as introduced in SVOO is applicable to any sparse-attention method, although it may give better results for some than others. While this has not been possible due to budget constraints, it would be interesting to evaluate how SVOO with offline calibration performs in our setup. Luo et al. appear to find that it provides a meaningful but limited improvement.

\subsection{DINOv2}

We sample the first 12 of the DIV8K \cite{9021973} images with a random generation seed of 42. 
% switched from 0 to get more diverse images, done before evaluation
These are shown in Appendix~\ref{div8klarge}. 
The images are square cropped and downscaled from their native to the evaluation resolution. We measure at three resolutions: 2072 by 2072 pixels, resulting in 21,904 tokens, 3500 by 3500 pixels, resulting in 62,500 tokens, and 5306 by 5306 pixels, meaning 143,641 tokens. We perform measurements on a Nvidia H100 GPU. Approximate attention is not applied to the \texttt{CLS}-token. All top-$k$ methods are evaluated at points where they attend to each of \{0.05, 0.1, 0.2, 0.4, 0.6, 0.8, 1.0\} of all keys, while adaptive methods are evaluated at each of \{0.6, 0.8, 0.9, 0.95, 0.99, 1\} as targeted attention-mass recall. \\

\subsection{TabPFN-3}
We exclude the \texttt{scikit-learn} processing and clock only the model backbone, as the preprocessing had high run-to-run latency variance. We also exclude test-to-train processing, as this is not sparsified, instead only timing train-to-train processing (including caching). As a warm-up we did three forward passes on each dataset. While ClusterAttention does not have overhead on the first run on a new tensor size (when run without CUDA-graphs as was done here), both the TabPFN-3 model and SVOO did have this property, so a universal warm-up could not be used. The top-$k$ methods sweep through including \{10\%, 40\%, 80\%, 100\%\} of all keys (default ClusterAttention adds 1\% and 5\%), while the adaptive methods sweep through targeting \{80\%, 90\%, 100\%\} of attention-mass. For the sparse methods, we evaluated the QASSMax function \cite{qu2026tabiclv2betterfasterscalable} (which adjusts softmax sharpness to the dataset length) with the length of the full dataset as parameter, instead of the actual number of keys attended to. We chose to do this as the distribution of keys is likely affected by the number of data points, so this parameter is not just tuning for the raw number of data points. It is also difficult to select for adaptive methods, where the number of keys attended to is not known beforehand. All methods except for the native dense need to transpose the data from (batch, sample, head, dimension) to (batch, head, sample, dimension) each forward pass, meaning that these methods carry a small performance penalty that could be removed by making the methods handle this layout. 

\subsection{Wan 2.1-T2V-14B}
We run the model on a Nvidia H200 GPU to match the configuration by Luo et al. for SVOO. Matching the SVOO paper, we use full density for the first layer out of the 40 in the model, and the first 10 out of 50 diffusion steps. While we did not perceive a visual degradation from running fully sparse, the inserted dense computations did decrease the deviation in generated content, making the metrics based on visual comparison more meaningful. As warm-up, we ran the model for one dense diffusion-step and two (optionally, not for the dense baseline) sparse steps. We ran one generation per prompt, all using random generation seed 0. 

\subsection{Latency scaling}
For variable input-length use cases, such as tabular models, there is slightly more constant-term overhead in the current implementation of the clustering than in Figure~\ref{scaling}. For fixed token-count cases, such as image models as shown here, it is removed using CUDA-graphs. The extra overhead is only noticeable at smaller token counts.\\

\newpage
\section{Evaluation images}
\label{div8klarge}
\begin{figure}[h]
    \centering
    \includegraphics[width=0.8\linewidth]{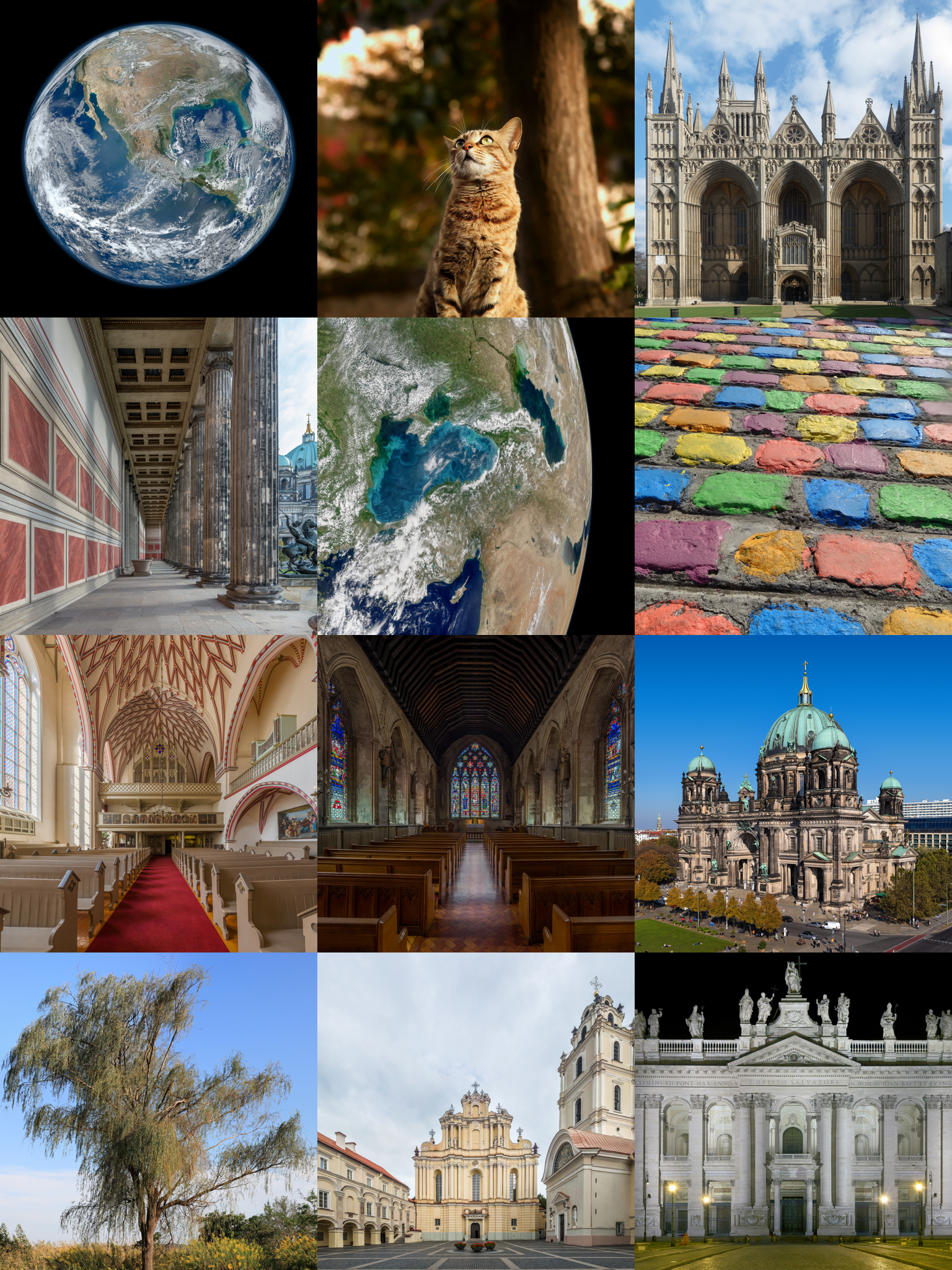}
    \caption{12 sampled high-resolution images from DIV8K \cite{9021973} after square-cropping, used for evaluation on DINOv2-L.}
    \label{div8klarge-collage}
\end{figure}

\newpage
\section{Adaptive methods on DINOv2}
\label{svoo-dino}

\begin{figure}[h]
  \centering
  \begin{subfigure}[b]{0.32\textwidth}
    \includegraphics[width=\textwidth]{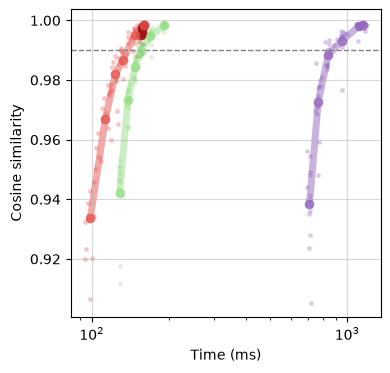}
    \caption{2072 by 2072 pixels.}
    \label{adaptive_small}
  \end{subfigure}
  \hfill
  \begin{subfigure}[b]{0.32\textwidth}
    \includegraphics[width=\textwidth]{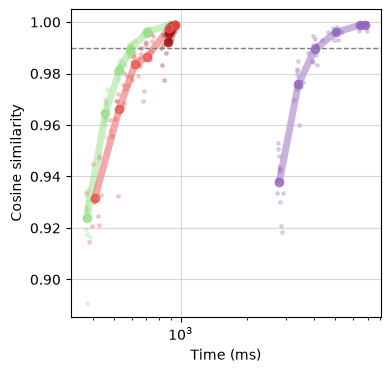}
    \caption{3500 by 3500 pixels.}
    \label{adaptive_medium}
  \end{subfigure}
  \hfill
  \begin{subfigure}[b]{0.32\textwidth}
    \includegraphics[width=\textwidth]{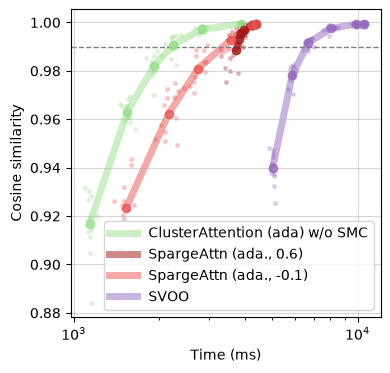}
    \caption{5306 by 5306 pixels.}
    \label{adaptive_large}
  \end{subfigure}
  \caption{Comparison between adaptive ClusterAttention and SVOO on DINOv2. Pareto front on latency vs. representation-distortion, measured in average cosine-similarity with dense.}
  \label{adaptive-dino-pareto}
\end{figure}

\section{TabPFN-3 Pareto sweeps}
\label{tabpfn-pareto-appendix}

\begin{figure}[!h]
    \centering
    \begin{subfigure}{0.3\textwidth}
        \includegraphics[width=\linewidth]{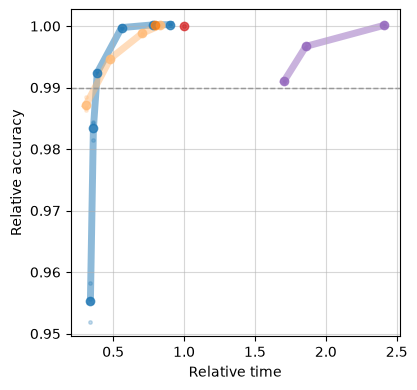}
        \caption{\textttlb{Rain\_in\_Australia} (93094 training rows).}
    \end{subfigure}
    \hfill
    \begin{subfigure}{0.3\textwidth}
        \includegraphics[width=\linewidth]{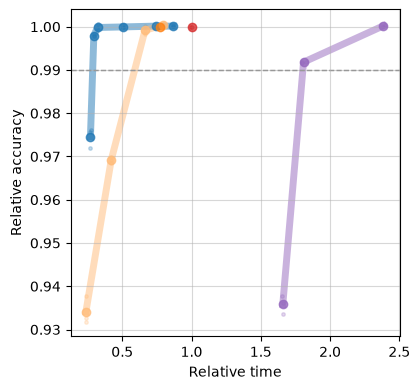}
        \caption{\textttlb{walking-activity} (95572 training rows).}
    \end{subfigure}
    \hfill
    \begin{subfigure}{0.3\textwidth}
        \includegraphics[width=\linewidth]{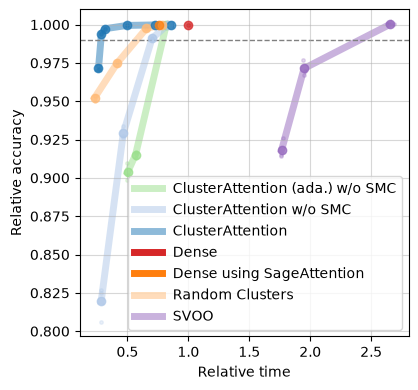}
        \caption{\textttlb{accelerometer} (97922 training rows).}
        \label{accelerometer}
    \end{subfigure}

    \begin{subfigure}{0.3\textwidth}
        \includegraphics[width=\linewidth]{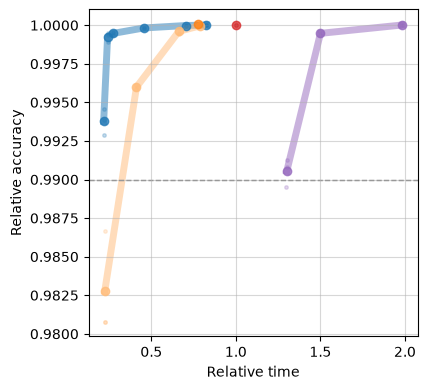}
        \caption{\textttlb{CDC\_Diabetes\_Health\_Indicators} (162355 training rows).}
    \end{subfigure}
    \hfill
    \begin{subfigure}{0.3\textwidth}
        \includegraphics[width=\linewidth]{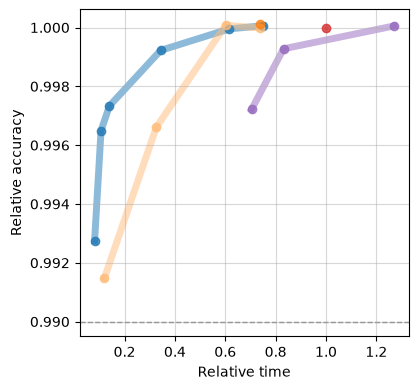}
        \caption{\textttlb{Data\_Science\_for\_Good\_Kiva\_Crowdfunding} (429571 training rows).}
    \end{subfigure}
    \hfill
    \begin{subfigure}{0.3\textwidth}
        \includegraphics[width=\linewidth]{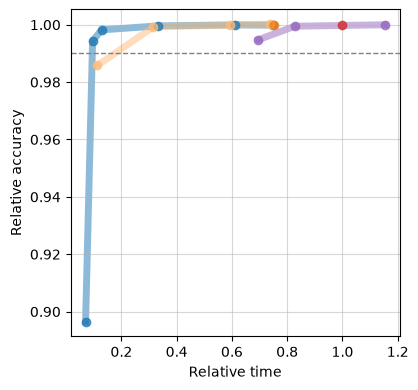}
        \caption{\textttlb{Smoking\_and\_Drinking\_Dataset\_with\_body\_signal} (634460 training rows).}
    \end{subfigure}

    \caption{Accuracy-latency Pareto front on TabPFN-3 for the 6 largest classification datasets from the TALENT dataset suite, with accuracy and latency normalized by the values with dense attention.}
    \label{tabpfn-paretos}
\end{figure}

\section{Other methods targeting the same setting as ClusterAttention}
\label{same-niche-methods}
\begin{figure}[h]
    \centering
    \includegraphics[width=0.6\textwidth]{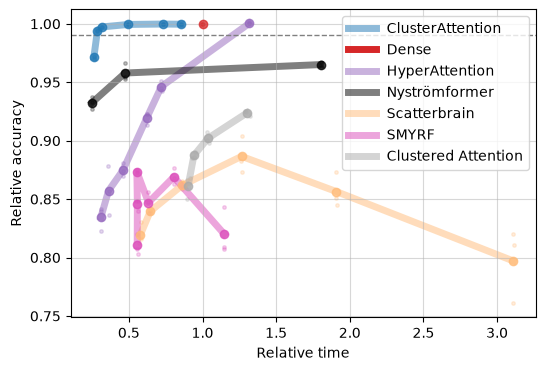}
    \caption{Comparison between ClusterAttention and baselines that target the setting of unstructured data and a single forward pass on TabPFN-3, on the dataset \texttt{accelerometer}, with the setup described in Appendix~\ref{experimental-setups}.}
  \label{same-niche}
\end{figure}

We test the methods that are described in Section~\ref{related-works} to target the setting of unstructured data and a single forward pass. We evaluate them on TabPFN-3, which falls into this setting, on the \texttt{accelerometer} dataset with 97922 training rows, chosen arbitrarily based on being the alphabetically first out of the datasets we use. As can be seen, the methods do not appear to provide speedups over dense attention at the operating point of 0.99 of dense accuracy chosen for TabPFN-3 in Subsection~\ref{tabpfn-results}. Arguably, the most competitive methods are the Nyströmformer \cite{xiong2021nystromformernystrombasedalgorithmapproximating} and HyperAttention \cite{han2023hyperattentionlongcontextattentionnearlinear}. Nyströmformer seems to plateau before reaching dense accuracy. Clustered Attention \cite{vyas2020fasttransformersclusteredattention} should trivially reach it if increasing the cluster-count up to the number of data points, or if increasing the top-$k$ to the same number. SMYRF \cite{daras2020smyrfefficientattentionusing} and Scatterbrain \cite{chen2021scatterbrain}, which here uses SMYRF and Performer \cite{choromanski2022rethinkingattentionperformers} internally, both appear to not strictly improve in quality with increased cluster-size. \\

\subsection{Configurations}
For all methods, the parameters we do not sweep are kept at the defaults of the reference implementations, with a few exceptions noted below. Furthermore, Scatterbrain, SMYRF, and Clustered Attention run with 32-bit floating point numbers, and we enable TF32 matrix multiplications for them to minimize the speed disadvantage from this on the Nvidia H100 GPU, though the difference was not large. 
HyperAttention and Nyströmformer natively run in the 16-bit brain floating point numbers that they are provided with by the model, although the pseudoinverse of the Nyströmformer is computed in 32-bit floating point numbers, as it is a precision sensitive operation.\\

For Scatterbrain, we use the combination of Performer and SMYRF and sweep the cluster size through $\{32, 64, 128, 256, 512, 1024\}$. We cap the number of Performer features, which in the reference configurations is set to match the cluster size, to $D\log D$, which is the general default for the Performer module in their implementation. To avoid going out of memory in its SMYRF part, we serialize over the attention heads.\\

For the Nyströmformer, we sweep the number of landmarks through $\{512, 1024, 2048\}$. Since the original segment-mean landmarks assume local structure, we instead sample the landmarks uniformly at random, and omit the residual convolution. We note that it is sensitive to too few or too many pseudoinverse-iterations, with the best results at the default of 6.\\

For HyperAttention, we sweep the parameter pair block size and sample size through $\{(6528, 256), \allowbreak(10880, 256), \allowbreak(19584, 256), (10880, 2048), \allowbreak(19584, 2048), \allowbreak(97920, 256)\}$. Since it does not handle sequence lengths that are not a multiple of the block size (other than by zero-padding them, which introduces a bias into the attention computation), we leave a remainder that is computed densely. The remainder queries attend to all keys, and the attention from the other queries to the remainder keys is merged with the HyperAttention output using the attention denominator. Since HyperAttention estimates the attention tail, i.e. the keys not attended to densely, this should not bias the output with regard to the remainder. We also select block sizes that leave a remainder of 2 on the \texttt{accelerometer} dataset, minimizing its impact on the output. \\

For SMYRF, we sweep the cluster size through $\{32, 64, 128, 256, 512, 1024\}$. We take the remaining parameters from the reference downstream NLP example. The method materializes a few large tensors, so to keep the GPU from going out-of-memory, we serialize over attention heads and hash rounds. The hash rounds are then merged using the attention denominators, which is exactly what the original implementation does. 
SMYRF requires the sequence length to be a multiple of the cluster size, so we pad to a multiple and mask as in the reference BERT integration. We use $-\infty$ instead of their $-10^4$ as the mask value, since we cannot be sure that norms stay small relative to it on another model. \\

For Clustered Attention, we use the improved variant with the default top-$k$ of 32, and sweep the number of clusters through $\{100, 250, 500, 1000\}$. 

%\newpage
\section{Ablations}
\label{ablations}
%Figure~\ref{ablation-transforms-topk} shows the performance at different resolutions for versions of ClusterAttention using top-$k$ assignment. 
\subsection{Transforms}
\label{transform-ablations}

\begin{figure}[h]
  \centering
  \begin{subfigure}[b]{0.3\textwidth}
    \includegraphics[width=\textwidth]{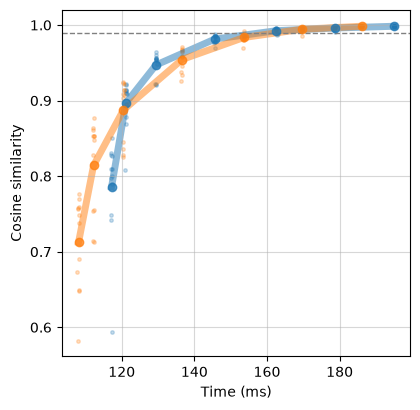}
    \caption{2072 by 2072 pixels.}
    \label{ablation_topk_small}
  \end{subfigure}
  \hfill
  \begin{subfigure}[b]{0.3\textwidth}
    \includegraphics[width=\textwidth]{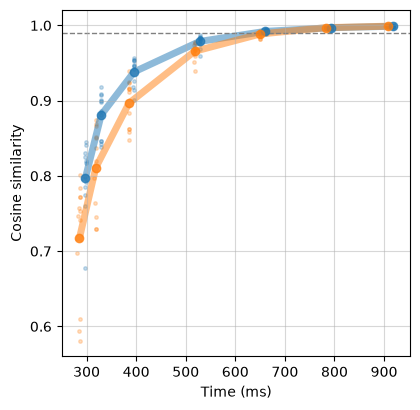}
    \caption{3500 by 3500 pixels.}
    \label{ablation_topk_medium}
  \end{subfigure}
  \hfill
  \begin{subfigure}[b]{0.3\textwidth}
    \includegraphics[width=\textwidth]{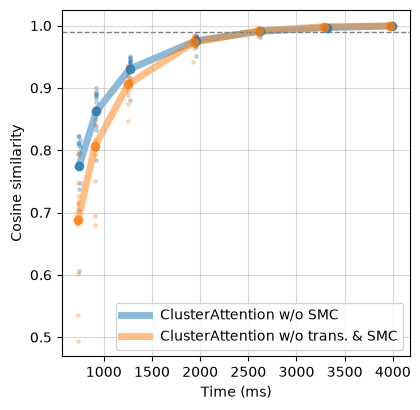}
    \caption{5306 by 5306 pixels.}
    \label{ablation_topk_large}
  \end{subfigure}
  \caption{Pareto front on latency vs. representation-distortion on DINOv2, for the top-$k$ methods without SMC. The dashed line represents 0.99 cosine similarity with dense.}
  \label{transform-ablation-topk}
\end{figure}

\begin{figure}[h]
  \centering
  \begin{subfigure}[b]{0.3\textwidth}
    \includegraphics[width=\textwidth]{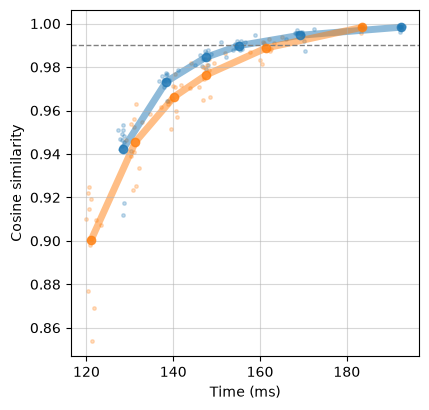}
    \caption{2072 by 2072 pixels.}
    \label{ablation_ada_small}
  \end{subfigure}
  \hfill
  \begin{subfigure}[b]{0.3\textwidth}
    \includegraphics[width=\textwidth]{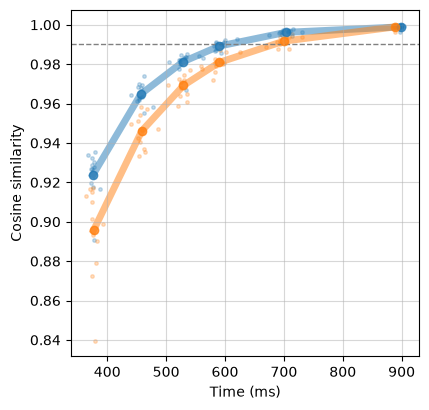}
    \caption{3500 by 3500 pixels.}
    \label{ablation_ada_medium}
  \end{subfigure}
  \hfill
  \begin{subfigure}[b]{0.3\textwidth}
    \includegraphics[width=\textwidth]{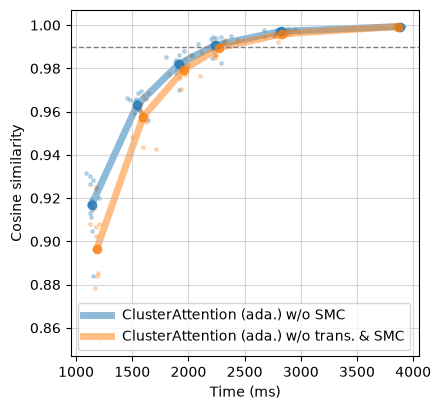}
    \caption{5306 by 5306 pixels.}
    \label{ablation_ada_large}
  \end{subfigure}
  \caption{Pareto front on latency vs. representation-distortion on DINOv2, for the adaptive methods without SMC. The dashed line represents 0.99 cosine similarity with dense.}
  \label{transform-ablation-ada}
\end{figure}
%\clearpage
\begin{figure}[h]
  \centering
  \begin{subfigure}[b]{0.3\textwidth}
    \includegraphics[width=\textwidth]{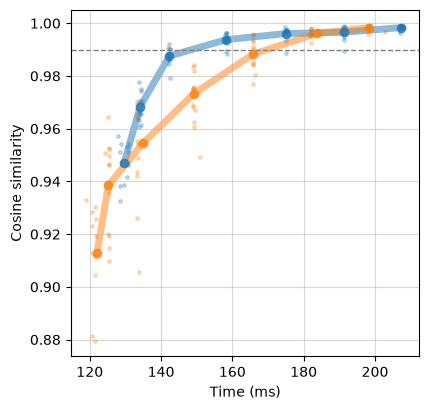}
    \caption{2072 by 2072 pixels.}
    \label{transform_ablation_topk_comp_small}
  \end{subfigure}
  \hfill
  \begin{subfigure}[b]{0.3\textwidth}
    \includegraphics[width=\textwidth]{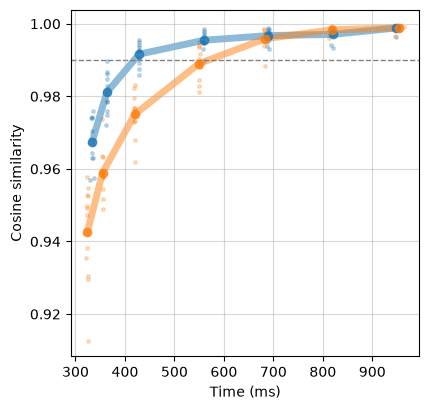}
    \caption{3500 by 3500 pixels.}
    \label{transform_ablation_topk_comp_medium}
  \end{subfigure}
  \hfill
  \begin{subfigure}[b]{0.3\textwidth}
    \includegraphics[width=\textwidth]{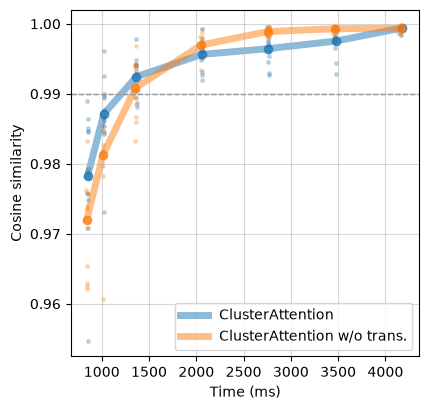}
    \caption{5306 by 5306 pixels.}
    \label{transform_ablation_topk_comp_large}
  \end{subfigure}
  \caption{Ablation of transforms. Pareto front on latency vs. representation-distortion on DINOv2, for the top-$k$ methods with SMC. The dashed line represents 0.99 cosine similarity with dense.}
  \label{transform-ablation-topk-comp}
\end{figure}

\begin{figure}[h]
  \centering
  \begin{subfigure}[b]{0.3\textwidth}
    \includegraphics[width=\textwidth]{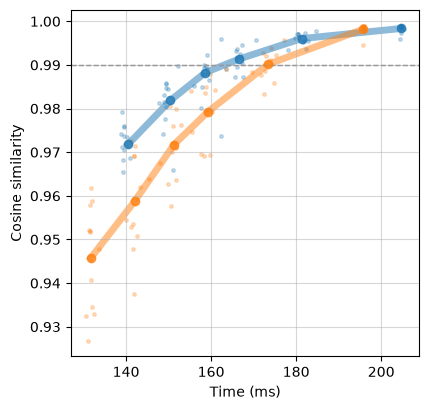}
    \caption{2072 by 2072 pixels.}
    \label{transform_ablation_ada_comp_small}
  \end{subfigure}
  \hfill
  \begin{subfigure}[b]{0.3\textwidth}
    \includegraphics[width=\textwidth]{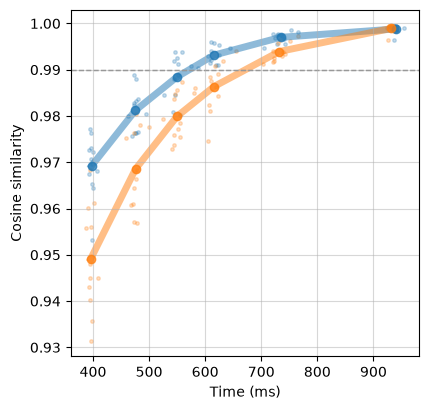}
    \caption{3500 by 3500 pixels.}
    \label{transform_ablation_ada_comp_medium}
  \end{subfigure}
  \hfill
  \begin{subfigure}[b]{0.3\textwidth}
    \includegraphics[width=\textwidth]{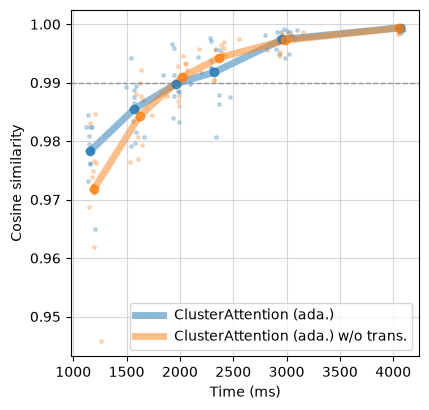}
    \caption{5306 by 5306 pixels.}
    \label{transform_ablation_ada_comp_large}
  \end{subfigure}
  \caption{Ablation of transforms. Pareto front on latency vs. representation-distortion on DINOv2, for the adaptive methods with SMC. The dashed line represents 0.99 cosine similarity with dense.}
  \label{transform-ablation-ada-comp}
\end{figure}

% \newpage
% \vfill
\clearpage
\subsection{Striped mean-compensation}
\label{compensation-ablation}

\begin{figure}[h]
  \centering
  \begin{subfigure}[b]{0.3\textwidth}
    \includegraphics[width=\textwidth]{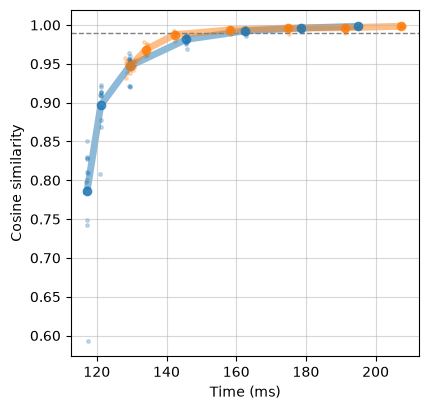}
    \caption{2072 by 2072 pixels.}
    \label{comp_ablation_topk_small}
  \end{subfigure}
  \hfill
  \begin{subfigure}[b]{0.3\textwidth}
    \includegraphics[width=\textwidth]{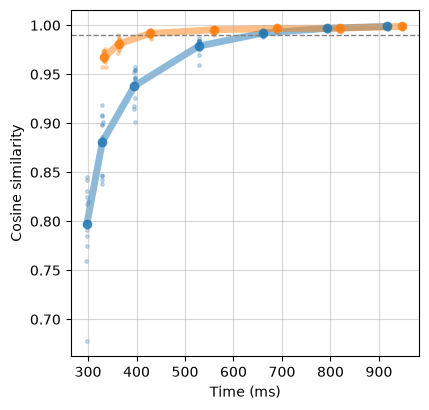}
    \caption{3500 by 3500 pixels.}
    \label{comp_ablation_topk_medium}
  \end{subfigure}
  \hfill
  \begin{subfigure}[b]{0.3\textwidth}
    \includegraphics[width=\textwidth]{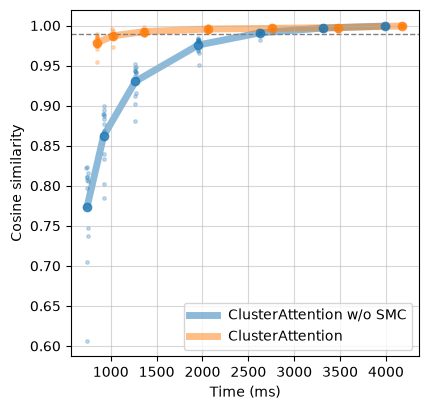}
    \caption{5306 by 5306 pixels.}
    \label{comp_ablation_topk_large}
  \end{subfigure}
  \caption{Ablation of compensation. Pareto front on latency vs. representation-distortion on DINOv2, for the top-$k$ methods. The dashed line represents 0.99 cosine similarity with dense.}
  \label{comp-ablation-topk}
\end{figure}

\begin{figure}[h]
  \centering
  \begin{subfigure}[b]{0.3\textwidth}
    \includegraphics[width=\textwidth]{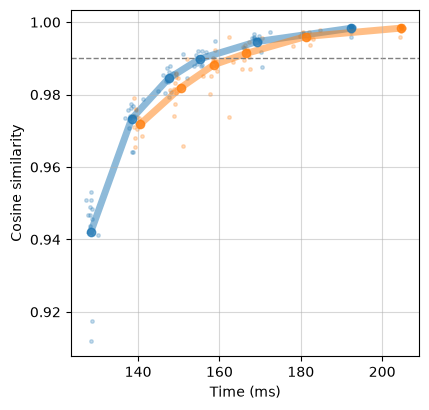}
    \caption{2072 by 2072 pixels.}
    \label{comp_ablation_ada_small}
  \end{subfigure}
  \hfill
  \begin{subfigure}[b]{0.3\textwidth}
    \includegraphics[width=\textwidth]{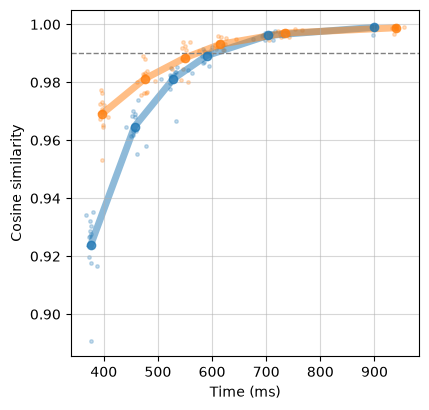}
    \caption{3500 by 3500 pixels.}
    \label{comp_ablation_ada_medium}
  \end{subfigure}
  \hfill
  \begin{subfigure}[b]{0.3\textwidth}
    \includegraphics[width=\textwidth]{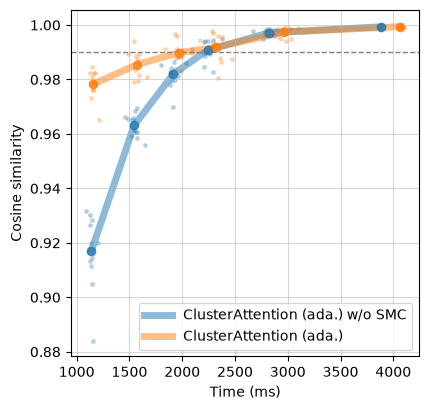}
    \caption{5306 by 5306 pixels.}
    \label{comp_ablation_ada_large}
  \end{subfigure}
  \caption{Ablation of compensation. Pareto front on latency vs. representation-distortion on DINOv2, for the adaptive methods. The dashed line represents 0.99 cosine similarity with dense.}
  \label{comp-ablation-ada}
\end{figure}

We find that in the top-$k$ case, the compensation provides a large quality improvement such that using compensation is Pareto-dominant except for at the highest sparsity at the smallest image size. In the adaptive case, the improvement is smaller, and mostly confined to the higher sparsities. At the lowest image size, the sweep without compensation Pareto-dominates the one with. We speculate that the mechanism behind why top-$k$ benefits more is the following. For a given computational budget, the adaptive method favors the queries with many clusters scoring on the higher end. But one mechanism causing clusters to score high, is that the keys and queries have small spread, giving the centroids larger magnitudes due to a smaller Jensen-effect. This causes the adaptive method to focus its effort on heads with small key and query variance. But for these heads, clusters are already a good summary, so the effort is misplaced. A confounding factor on this hypothesis is the difference in transforms for query-clustering.

\newpage
\subsection{Top-$k$ versus adaptive}
\label{selection-ablation}

\begin{figure}[h]
  \centering
  \begin{subfigure}[b]{0.3\textwidth}
    \includegraphics[width=\textwidth]{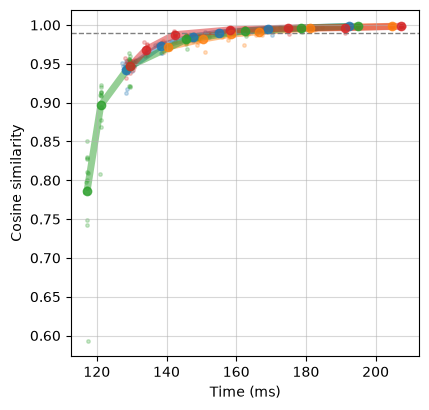}
    \caption{2072 by 2072 pixels.}
    \label{selection_ablation_small}
  \end{subfigure}
  \hfill
  \begin{subfigure}[b]{0.3\textwidth}
    \includegraphics[width=\textwidth]{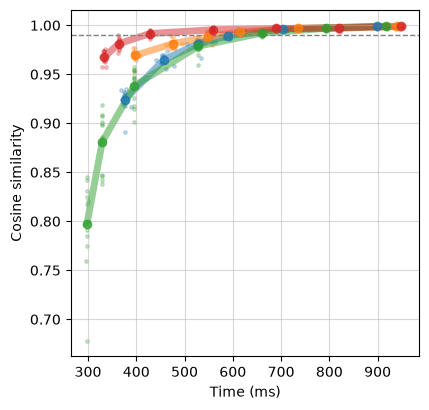}
    \caption{3500 by 3500 pixels.}
    \label{selection_ablation_medium}
  \end{subfigure}
  \hfill
  \begin{subfigure}[b]{0.3\textwidth}
    \includegraphics[width=\textwidth]{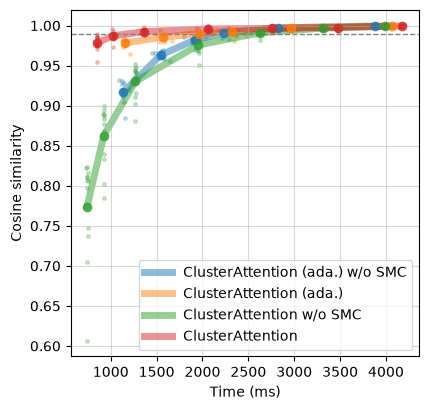}
    \caption{5306 by 5306 pixels.}
    \label{selection_ablation_large}
  \end{subfigure}
  \caption{Ablation of selection method. Pareto front on latency vs. representation-distortion on DINOv2. The dashed line represents 0.99 cosine similarity with dense.}
  \label{selection-ablation-pareto}
\end{figure}

%\newpage
\section{OpenSora 1.0 prompts}
\label{prompts}
1. A bustling city street at night, filled with the glow of car headlights and the ambient light of streetlights. The scene is a blur of motion, with cars speeding by and pedestrians navigating the crosswalks. The cityscape is a mix of towering buildings and illuminated signs, creating a vibrant and dynamic atmosphere. The perspective of the video is from a high angle, providing a bird's eye view of the street and its surroundings. The overall style of the video is dynamic and energetic, capturing the essence of urban life at night.\\\\
2. A detailed wooden toy ship with intricately carved masts and sails is seen gliding smoothly over a plush, blue carpet that mimics the waves of the sea. The ship's hull is painted a rich brown, with tiny windows. The carpet, soft and textured, provides a perfect backdrop, resembling an oceanic expanse. Surrounding the ship are various other toys and children's items, hinting at a playful environment. The scene captures the innocence and imagination of childhood, with the toy ship's journey symbolizing endless adventures in a whimsical, indoor setting.\\\\
3. A majestic beauty of a waterfall cascading down a cliff into a serene lake. The waterfall, with its powerful flow, is the central focus of the video. The surrounding landscape is lush and green, with trees and foliage adding to the natural beauty of the scene. The camera angle provides a bird's eye view of the waterfall, allowing viewers to appreciate the full height and grandeur of the waterfall. The video is a stunning representation of nature's power and beauty.\\\\
4. A serene night scene in a forested area. The first frame shows a tranquil lake reflecting the star-filled sky above. The second frame reveals a beautiful sunset, casting a warm glow over the landscape. The third frame showcases the night sky, filled with stars and a vibrant Milky Way galaxy. The video is a time-lapse, capturing the transition from day to night, with the lake and forest serving as a constant backdrop. The style of the video is naturalistic, emphasizing the beauty of the night sky and the peacefulness of the forest.\\\\
5. A serene underwater scene featuring a sea turtle swimming through a coral reef. The turtle, with its greenish-brown shell, is the main focus of the video, swimming gracefully towards the right side of the frame. The coral reef, teeming with life, is visible in the background, providing a vibrant and colorful backdrop to the turtle's journey. Several small fish, darting around the turtle, add a sense of movement and dynamism to the scene. The video is shot from a slightly elevated angle, providing a comprehensive view of the turtle's surroundings. The overall style of the video is calm and peaceful, capturing the beauty and tranquility of the underwater world.\\\\
6. A snowy forest landscape with a dirt road running through it. The road is flanked by trees covered in snow, and the ground is also covered in snow. The sun is shining, creating a bright and serene atmosphere. The road appears to be empty, and there are no people or animals visible in the video. The style of the video is a natural landscape shot, with a focus on the beauty of the snowy forest and the peacefulness of the road.\\\\
7. A soaring drone footage captures the majestic beauty of a coastal cliff, its red and yellow stratified rock faces rich in color and against the vibrant turquoise of the sea. Seabirds can be seen taking flight around the cliff's precipices. As the drone slowly moves from different angles, the changing sunlight casts shifting shadows that highlight the rugged textures of the cliff and the surrounding calm sea. The water gently laps at the rock base and the greenery that clings to the top of the cliff, and the scene gives a sense of peaceful isolation at the fringes of the ocean. The video captures the essence of pristine natural beauty untouched by human structures.\\\\
8. A vibrant scene of a snowy mountain landscape. The sky is filled with a multitude of colorful hot air balloons, each floating at different heights, creating a dynamic and lively atmosphere. The balloons are scattered across the sky, some closer to the viewer, others further away, adding depth to the scene.  Below, the mountainous terrain is blanketed in a thick layer of snow, with a few patches of bare earth visible here and there. The snow-covered mountains provide a stark contrast to the colorful balloons, enhancing the visual appeal of the scene.  In the foreground, a few cars can be seen driving along a winding road that cuts through the mountains. The cars are small compared to the vastness of the landscape, emphasizing the grandeur of the surroundings.  The overall style of the video is a mix of adventure and tranquility, with the hot air balloons adding a touch of whimsy to the otherwise serene mountain landscape. The video is likely shot during the day, as the lighting is bright and even, casting soft shadows on the snow-covered mountains.\\
9. A vibrant underwater scene. A group of blue fish, with yellow fins, are swimming around a coral reef. The coral reef is a mix of brown and green, providing a natural habitat for the fish. The water is a deep blue, indicating a depth of around 30 feet. The fish are swimming in a circular pattern around the coral reef, indicating a sense of motion and activity. The overall scene is a beautiful representation of marine life.\\\\
10. The camera follows behind a white vintage SUV with a black roof rack as it speeds up a steep dirt road surrounded by pine trees on a steep mountain slope, dust kicks up from its tires, the sunlight shines on the SUV as it speeds along the dirt road, casting a warm glow over the scene. The dirt road curves gently into the distance, with no other cars or vehicles in sight. The trees on either side of the road are redwoods, with patches of greenery scattered throughout. The car is seen from the rear following the curve with ease, making it seem as if it is on a rugged drive through the rugged terrain. The dirt road itself is surrounded by steep hills and mountains, with a clear blue sky above with wispy clouds.\\\\
11. The dynamic movement of tall, wispy grasses swaying in the wind. The sky above is filled with clouds, creating a dramatic backdrop. The sunlight pierces through the clouds, casting a warm glow on the scene. The grasses are a mix of green and brown, indicating a change in seasons. The overall style of the video is naturalistic, capturing the beauty of the landscape in a realistic manner. The focus is on the grasses and their movement, with the sky serving as a secondary element. The video does not contain any human or animal elements.\\\\
12. The vibrant beauty of a sunflower field. The sunflowers, with their bright yellow petals and dark brown centers, are in full bloom, creating a stunning contrast against the green leaves and stems. The sunflowers are arranged in neat rows, creating a sense of order and symmetry. The sun is shining brightly, casting a warm glow on the flowers and highlighting their intricate details. The video is shot from a low angle, looking up at the sunflowers, which adds a sense of grandeur and awe to the scene. The sunflowers are the main focus of the video, with no other objects or people present. The video is a celebration of nature's beauty and the simple joy of a sunny day in the countryside.

\end{document}